\documentclass{article}
\usepackage[preprint]{neurips_2021}

\usepackage[utf8]{inputenc} 
\usepackage[T1]{fontenc}    
\usepackage{hyperref}       
\usepackage{url}            
\usepackage{booktabs}       
\usepackage{amsfonts}       
\usepackage{nicefrac}       
\usepackage{microtype}      
\usepackage{natbib}
\usepackage{soul}

\newcommand\smallO{\scriptstyle\scriptscriptstyle\mathcal{O}}
\newcommand\smallM{\scriptstyle\scriptscriptstyle\mathcal{M}}
    
\newcommand{\review}[1]{\textcolor{black}{#1}}

\newcommand{\x}{{\bm{x}}}
\newcommand{\z}{{\bm{z}}}

\usepackage{subcaption}
\usepackage{wrapfig}
\usepackage{bm}
\usepackage{amsmath}
\usepackage{amsfonts}
\usepackage{multirow}
\usepackage{amssymb}
\usepackage{pifont}
\usepackage[dvipsnames]{xcolor}
\usepackage{enumitem}
\usepackage[pdftex]{graphicx}
\newcommand{\ignore}[1]{}

\newcommand{\BB}[1]{$\normalsize \textbf{#1}$}

\title{Discriminative Multimodal Learning via \\ Conditional Priors in Generative Models}

\author{\normalfont Rogelio A. Mancisidor$^{a,\ast}$ \\ \href{mailto:rogelio.a.mancisidor@bi.no}{rogelio.a.mancisidor@bi.no}
\and Michael Kampffmeyer $^{b,c}$ \\
\href{mailto:michael.c.kampffmeyer@uit.no}{michael.c.kampffmeyer@uit.no}
\and Kjersti Aas $^c$ \\
\href{mailto:kjersti@nr.no}{kjersti@nr.no}
\and Robert Jenssen $^{b,c}$ \\
\href{mailto:robert.jenssen@uit.no}{robert.jenssen@uit.no}
\\
\\
\normalsize{$^{a}$ Department of Data Science and Analytics,}\\
\normalsize{ BI Norwegian Business School, Nydalsveien 37, 0484 Oslo}\\
\normalsize{$^{b}$ Department of Physics and Technology, Faculty of Science and Technology,}\\
\normalsize{UiT The Arctic University of Norway, Hansine Hansens veg 18, 9037 Troms{\o}}\\
\normalsize{$^{c}$Norwegian Computing Center, P.O. Box 114 Blindern Oslo}\\
\\
\normalsize{$^\ast$Corresponding author}
\
}

\begin{document}

\maketitle

\begin{abstract}
Deep generative models with latent variables have been used lately to learn joint representations and generative processes from multi-modal data. These two learning mechanisms can, however, conflict with each other and representations can fail to embed information on the data modalities. This research studies the realistic scenario in which all modalities and class labels are available for model training, but where some modalities and labels required for downstream tasks are missing. We show, in this scenario, that the variational lower bound limits mutual information between joint representations and missing modalities. We, to counteract these problems, introduce a novel conditional multi-modal discriminative model that uses an informative prior distribution and optimizes a likelihood-free objective function that maximizes mutual information between joint representations and missing modalities. Extensive experimentation demonstrates the benefits of our proposed model, empirical results show that our model achieves state-of-the-art results in representative problems such as downstream classification, acoustic inversion, and image and annotation generation.
\end{abstract}

\section{Introduction}\label{sec_intro}
Different measurement-modalities $\x_1,\x_2,\cdots,\x_m$ of an object are used in multi-modal learning to learn $\z$, a joint representation which captures information from all modalities. $\z$ can be used for clustering, active and transfer learning, or, where class labels $y$ are available, downstream classification. 

Deep neural networks (DNNs) and deep generative models (DGMs) with latent representations have been used in multi-modal learning \citep{andrew2013deep,wang2015deep,wang2016deep,suzuki2016joint,du2018semi,wu2018multimodal,du2019discriminative,shi2019variational,sutter2020multimodal,sutter2021generalized}. DGMs learn joint latent representations using variational approximations of the posterior distribution, and learn generative models for data modalities \review{by optimizing a variational lower bound on the log-likelihood of the data}. These two mechanisms can, however, conflict with each other. Generative models may focus on generating modalities without using the joint latent representation. Therefore, the posterior distribution for the joint representation fails to embed information on the modalities, collapsing into a non-informative prior distribution. This is called posterior collapse in the uni-modal domain \citep{dieng_avoiding_2019,lucas_dont_2019}, and harms the performance of downstream tasks based on joint representations, e.g. classification or generation of missing modalities.  

\review{There are different applications where data comes from different channels, e.g., tuples of images and annotations or acoustic and articulatory measurements. However, not all observations come in tuples, because annotating images or measuring articulatory movements can e.g. be costly \citep{sutter2020multimodal,sutter2021generalized} and take time to be generated. Hence, we are interested in modeling conditional distributions that are able to learn multi-modal latent representations, which can then be used to generate the missing modalities. Learning such latent representations is important, because we can capture relationships between modalities that are valuable for generative and discriminative downstream tasks. Towards this goal, we introduce a conditional multi-modal discriminative (CMMD) model that works in the aforementioned scenarios, where all modalities and class labels are available for model training, but where some modalities and class labels required for downstream tasks are missing.} We show, in this scenario, that the variational lower bound limits mutual information (MI) between multi-modal representations and missing modalities. To counteract this limitation and to circumvent posterior collapse, we introduce a novel likelihood-free objective function that optimizes MI and also introduce a prior distribution for joint representations that is conditioned on the available modalities. 

We show, through extensive experimentation, that our proposed CMMD model does not suffer from multi-modal posterior collapse. We also show that its joint representations embed information from multiple data modalities, which is useful to downstream tasks. We have benchmarked different multi-modal learning models across different representative domains, e.g. image-to-image, acoustic-to-articulatory, image-to-annotation, and text-to-image. The empirical results from this show that CMMD achieves state-of-the-art results in downstream classification and in the generation of missing modalities at test time. 

Our main contributions are:
\begin{itemize}
    \item A new objective function that counteracts the restriction on MI between joint representations and the missing modalities 
    \item A generative process that generates missing modalities at test time using a conditional prior\footnote{Conditional priors in variational autoencoders were introduced in \cite{sohn2015learning}. However, the focus was on the reconstruction of output data based on (always available) input data. On the other hand, our model uses conditional priors to generate latent representations in scenarios with missing data modalities. Hence, the prior distribution is modulated by the available modalities at test time and generates more informative representations than isotropic Gaussian priors.}
    \item Insights into the effect of posterior collapse in downstream classification and in the generative process in multi-modal learning.
\end{itemize}

\section{Multimodal Learning}\label{sec_relatedwork}
We use a common notation. Data modalities are represented by $\x$ and distinguished by a subscript. Joint latent representations are denoted by $\z$. In the following we provide an overview of the relevant multi-modal learning models to this work. See \cite{guo2019deep} for a comprehensive review.

\paragraph{Deep Neural Networks}
Deep canonical correlation analysis \citep{andrew2013deep} couples deep neural networks with canonical correlation analysis \citep{Hotelling1936} to train neural networks $f(\cdot)$ and $g(\cdot)$ such that they can maximize the correlation $\rho(f(\x_1),g(\x_2))$ between views (modalities) $\x_1$ and $\x_2$. DCCA (Deep Canonical Correlation Analysis) not only handles non-linearities, but also captures high-level data abstractions in each of the multiple hidden layers. Its objective function is, however, a function of the entire data set and therefore does not scale to large data sets. To overcome this limitation, \cite{wang2015deep} developed the deep canonically correlated autoencoder (DCCAE), which is optimized using stochastic gradient descent. DCCAE also introduced reconstruction neural networks for the data modalities, which minimized their reconstruction error. This is in addition to maximizing the canonical correlation between the learned representations. 

\paragraph{Variational Inference}
A problem with DCCAE is that the canonical correlation term in its objective function dominates the optimization procedure \citep{wang2015deep}. The reconstruction of the modalities is therefore poor. \cite{wang2016deep} therefore developed a variational CCA (VCCA) model to overcome this problem. VCCA uses variational inference and deep generative models to generate latent representations of the input modalities. 

\cite{du2019discriminative} proposed DMDGM, a supervised extension of VCCA that combines multi-modal learning and classification in a unified framework. The classification in DMDGM uses available views and not joint representations. DMDGM is, however, not the only model that addresses classification in a unified objective function. \cite{du2018semi} developed a semi-supervised deep generative model for missing modalities, the latent variable being shared across modalities. They also modeled the inference process as a Gaussian mixture model (GMM). Modeling the inference process as a GMM, however, harms the tightness of the lower bound, as the entropy of a GMM is intractable.

The model presented by \cite{vedantam2017generative} focuses on cross-modality generation, using the product of experts (PoE) in the factorization of posterior inference distributions. \cite{wu2018multimodal} similarly introduced MVAE, which assumes that the posterior distribution is proportional to the product of individual conditional posteriors $p(\z|\x_1) \cdots p(\z|\x_n)$ normalized by the prior distribution $p(\z)$. The joint posterior distribution is therefore also a PoE. \cite{shi2019variational}, through applying a similar approach, used a mixture of experts (MoE) to develop MMVAE, the generative process of the model allowing conditioning modalities and generation modalities to be interchangeable. MoE and PoE provide elegant ways of cross-generation. The linear combination of marginal distributions, however, learn joint representations that might not be useful for downstream classification (see Section \ref{sec_vimodels}). \cite{sutter2021generalized} show that the MVAE models the joint posterior distribution as a geometric mean, while MMVAE models it as an arithmetic mean. Further, they generalize these two approaches in a Mixture-of-Products-of-Experts (MoPoE) VAE, which approximates the joint posterior of all subsets of modalities. \review{It is noteworthy that MVAE, MMVAE, and MoPoE approximate the joint posterior distribution, conditioned on all modalities, as a function of unimodal posterior distributions. Such a modeling approach can deal with any combination of missing modalities simultaneously and, therefore, cross-modal generation can be done in any direction efficiently. However, none of these models are discriminative by nature and, as a consequence, can only deal with discriminative tasks in a two-steps fashion. CMMD is also able to model any combination of missing modalities, but one at a time. On the other hand, generative and discriminative models are trained end-to-end in the CMMD model.}

The most recent multi-modal learning research has focused on different ways of learning flexible joint representations that are useful in cross-modality generation. For example,  \cite{theodoridis2020cross} describe the learning of joint representation by introducing a cross-modal alignment of the latent spaces by minimizing Wasserstein distances; \cite{nedelkoski2020learning} couple normalizing flows and MVAE to learn more expressive representations; \cite{liu2021variational} propose a variational information bottleneck lower bound to force the encoder to discard irrelevant information, keeping only relevant information to generate one modality. \cite{chen2022multimodal} use generative adversarial networks to simultaneously align the different encoder distributions with the joint decoder distribution. None of these new methods, however, have been developed for downstream classification with missing modalities. \review{\cite{javaloy2022mitigating} focus on learning encoders and decoders that are impartial to the unimodal posterior distributions that generate latent representations. To achieve such impartial optimization (IO), the authors propose a novel optimization technique that modifies the gradients of each modality and, as a result, does not neglect the optimization of any specific modality.}

\cite{abrol2020improving} introduced a uni-modal method that uses, as in our proposed model, conditional priors to generate a discrete mixture of representations in the prior space. These are considered to be local latent variables. Continuous variables in the posterior distribution are  considered to be global. Local and global variables are, for supervised data, aligned using maximum mean discrepancy \citep{gretton2007kernel}, which optimizes the mutual information of global latent variables and input data. However, our proposed CMMD model focuses, instead, on multi-modal data and uses conditional priors to generate representations when some modalities are missing. Further, its objective function arises from the restriction imposed by the Kullback-Leibler divergence in the evidence lower bound on mutual information. 

\section{Methods}\label{sec_model}
\subsection{Evidence Lower Bound}
We have access to labeled multi-modal data $(\x_{\smallO},\x_{\smallM},y)$ during training. $\x_{\smallO}=(\x_1,\cdots,\x_n)$ are \textit{n} modalities that are always available and $\x_{\smallM}=(\x_{n+1},\cdots,\x_{n+m})$ are \textit{m} modalities that are missing at test time\footnote{Hence, subscripts in $\x_{\smallO}$ and $\x_{\smallM}$ indicate whether modalities are \textbf{o}bserved or \textbf{m}issing at test time.}. Only $\x_{\smallO}$ is therefore available for downstream tasks, the label $y$ and $\x_{\smallM}$ both  missing. Our proposed model at test time generates latent representations, using a prior distribution $p(\z|\x_{\smallO})$ conditioned on the observed modalities. Latent representations $\z \sim p(\z|\x_{\smallO})$ are furthermore used in both the generative process $p(\x_{\smallM}|\x_{\smallO},\z)$ and in the classifier model $p(y|\z)$. This encourages the model to learn useful representations for classification, and to generate missing modalities at test time. 

The joint distribution in our proposed model is, in this scenario, given by $p(\x_{\smallM},y,\z|\x_{\smallO})=p(\x_{\smallM}|\x_{\smallO},\z)p(y|\z)p(\z|\x_{\smallO})$, where $p(\z|\x_{\smallO})$ is a prior distribution conditioned on the always available modalities, $p(\x_{\smallM}|\x_{\smallO},\z)$ is the generative process for the missing modalities at test time, and $p(y|\z)$ is the density function for class labels. Note that the posterior distribution $p(\z|\x_{\smallO},\x_{\smallM},y)$, the joint latent representation that we want to learn, requires a marginal distribution that is not available in closed form. We therefore approximate the true posterior distribution $p(\z|\x_{\smallO},\x_{\smallM},y)$ using the parametric model, or encoder distribution, $q(\z|\x_{\smallO},\x_{\smallM},y)$. 

The evidence lower bound (ELBO) $\mathcal{L}(\x_{\smallM},\x_{\smallO},y)$ of our proposed model is therefore
\begin{align}
    \log p(\x_{\smallM},y|\x_{\smallO})
                      \geq& \mathbb{E}_{q(\z|\x_{\smallO},\x_{\smallM},y)}\Big[\log \frac{p(\x_{\smallM},y,\z|\x_{\smallO})}{q(\z|\x_{\smallO},\x_{\smallM},y)}\Big] \nonumber \\
                      \equiv& \mathcal{L}(\x_{\smallO},\x_{\smallM},y),
\label{eq_lowerbound}
\end{align}
the inequality being a result of the concavity of log and Jensen's inequality. \review{See Appendix \ref{ap_lowerbound} for details.}

\subsection{Maximizing Mutual Information}
We can, in principle, optimize Eq. \ref{eq_lowerbound} using the stochastic variational gradient Bayes (SVGB) algorithm \citep{kingma2013auto}. Eq. \ref{eq_lowerbound} does, however, include an average Kullback-Leibler divergence that is an upper bound on the conditional mutual information between $\z$ and $\x_{\smallM}$ (see Appendix B), i.e. 
\begin{align}
\mathbb{E}_{p(\x_{\smallM},\x_{\smallO},y)}[KL[q(\z|\x_{\smallO},\x_{\smallM},y)||p(\z|\x_{\smallO})] \geq I(\x_{\smallM},\z|\x_{\smallO}).
\label{eq_MI_bound}
\end{align}

The latent representations may therefore, as a consequence of this, fail to encode information about $\x_{\smallM}$, which is equivalent to generating $\x_{\smallM}$ based on the prior $p(\z|\x_{\smallO})$. This problem is called \textit{posterior collapse} in the one-modality literature \citep{dieng_avoiding_2019,lucas_dont_2019}. It occurs when the variational posterior distribution matches the prior. We therefore introduced a conditional mutual information term $(1-\omega)I(\x_{\smallM},\z|\x_{\smallO})$ in Eq. \ref{eq_lowerbound} to counteract this problem, $\omega \in [0,1]$ weighting the optimization on the mutual information term. The following likelihood-free objective function for a single data point is therefore obtained\footnote{We write the objective function for a single data point to improve readability. The outer expectation in the objective function for the average conditional log-likelihood is approximated with the empirical data distribution.} 
\begin{align}
\mathcal{J}(\x_{\smallO},\x_{\smallM},y) =&\mathbb{E}_{q_{(\z|\x_{\smallO},\x_{\smallM},y)}}[\log p(\x_{\smallM}|\x_{\smallO},\z) + \log p(y|\z) + \log p(\z|\x_{\smallO}) \nonumber \\
-& \log q(\z|\x_{\smallO},\x_{\smallM},y)] + (1-\omega) I(\x_{\smallM},\z) \nonumber \\
=& \mathbb{E}_{q_{(\z|\x_{\smallO},\x_{\smallM},y)}}[\log p(\x_{\smallM}|\x_{\smallO},\z)+\log p(y|\z)] \nonumber \\
-& \omega KL[q(\z|\x_{\smallO},\x_{\smallM},y)||p(\z|\x_{\smallO})] 
- (1-\omega) KL[q(\z|\x_{\smallO})||p(\z|\x_{\smallO})],
\label{eq_elbo_2kl}
\end{align}
where the last divergence term is called the marginal KL divergence \citep{hoffman2016elbo}. The full derivation for Eq. \ref{eq_elbo_2kl} is given in Appendix A.

The first KL divergence term in Eq. \ref{eq_elbo_2kl} has an analytical solution. The second KL divergence is intractable due to the marginal distribution $q(\z|\x_{\smallO})$. It can, however, be replaced by any strict divergence term \citep{zhao2017infovae}, e.g. maximum mean discrepancy divergence (MMD) \citep{gretton2007kernel}. We select the squared population MMD since it encourages the average posterior distribution to match the whole prior, which is
\begin{align}
    \text{\footnotesize MMD}[\mathcal{F},p,q] = \mathbb{E}_{p(x,x')}[k(x,x')]-2\mathbb{E}_{p(x),q(z)}[k(x,z)] + \mathbb{E}_{q(z,z')}[k(z,z')].
    \label{eq_MMD}
\end{align}
Here $\mathcal{F}$ is a unit ball in a universal reproducing kernel Hilbert space $\mathcal{H}$, $p$ and $q$ are Borel probability measures, and $k(\cdot,\cdot)$ is a universal kernel. We use a Gaussian kernel in our proposed model. Finally, the objective function for a single data point therefore becomes
\begin{align}
\mathcal{J}(\x_{\smallM},\x_{\smallO},y) =& \mathbb{E}_{q(\z|\x_{\smallO},\x_{\smallM})}[\log p(\x_{\smallM}|\x_{\smallO},\z) + \alpha \log p(y|\z)] \nonumber \\
-& \omega  KL[q(\z|\x_{\smallO},\x_{\smallM},y)||p(\z|\x_{\smallO})] - (1-\omega)  \lambda \text{\footnotesize MMD}[q(\z|\x_{\smallO}),p(\z|\x_{\smallO})],
\label{eq_convex_elbo}
\end{align}
where $\lambda$ counteracts the loss imbalance between the $\mathcal{X}_2$ and $\mathcal{Z}$ spaces and $\alpha$ controls the importance of the classification loss in the objective function.

\paragraph{Effect of $\bm{\omega}$ on the objective function:} \review{The first (term-by-term) KL divergence in Eq. \ref{eq_convex_elbo} regularizes each posterior distribution towards its prior and is minimized when $q_i(\z|\x_{\smallO}^{i},\x_{\smallM}^{i},y^{i}) = p_i(\z|\x_{\smallO}^{i})$ for all $i$. The marginal MMD divergence, on the other hand, regularizes an \textit{average posterior} distribution $q(\z|\x_{\smallO})=1/N \sum_i q(\z|\x_{\smallO},\x_{\smallM}^i,y^i)$ towards the prior distribution, without sacrificing model power \citep{hoffman2016elbo}. \cite{makhzani2015adversarial} show that the term-by-term KL divergence simply encourages the average posterior distribution to match the modes of the prior $p(\z|\x_{\smallO})$. However, the MMD term in Eq. \ref{eq_convex_elbo} encourages the average posterior distribution to match the whole prior, giving an effect similar to the adversarial training proposed by \cite{makhzani2015adversarial}. Furthermore, setting the marginal MMD divergence to 0 may lead to  representations from the prior that are useless for sculpting latent representations \citep{hoffman2016elbo}. Setting the term-by-term KL divergence to 0 also implies that the joint posterior representation is independent of the modality $\x_{\smallM}$. Our proposed objective function therefore offers an elegant way of trading-off these effects through the $\omega$ parameter, recovering the variational lower bound for $\omega=1$ and, for $1 > \omega \geq 0$, optimizing mutual information. The optimal $\omega$ value, as can be seen in Sections \ref{sec_mm}-\ref{sec_ai_ag} and \ref{sec_ablation}, is specific to the learning task and, therefore, must be found by cross-validation.}

CMMD finally assumes the following density functions for the prior distribution, the classifier, and the encoder
\begin{align*}
    p(\z|\x_{\smallO}) &\sim \mathcal{N}(\bm{\mu}=f_{\bm{\theta}}(\x_{\smallO}),\bm{\sigma}^2=f_{\bm{\theta}}(\x_{\smallO})), \\
    p(y|\z) &\sim \text{\footnotesize Cat} (\pi_{y|\z}=f_{\bm{\theta}}(\z)), \\
    q(\z|\x_{\smallO},\x_{\smallM},y) &\sim \mathcal{N}(\bm{\mu}=f_{\bm{\phi}}(\x_{\smallO},\x_{\smallM},y),\bm{\sigma}^2=f_{\bm{\phi}}(\x_{\smallO},\x_{\smallM},y)).
    \label{eq_distributions}
\end{align*}
The decoder network is parametrized as
\begin{align}
    p(\x_{\smallM}|\x_{\smallO},\z) &\sim \mathcal{N}(\bm{\mu}=f_{\bm{\theta}}(\x_{\smallO},\z), \bm{\sigma}^2=f_{\bm{\theta}}(\x_{\smallO},\z)),  \nonumber \\
\text{or} \nonumber \\
p(\x_{\smallM}|\x_{\smallO},\z) &\sim \text{\footnotesize Bernoulli}(\bm{p}=f_{\bm{\theta}}(\x_{\smallO},\z)), 
\end{align}
where $\mathcal{N}$ and Cat denote the Gaussian and multinomial distributions respectively, and where $f(\cdot)$ is a multilayer perceptron (MLP) network \citep{rumelhart1985learning}. This means that the density parameters $\bm{\mu}$, $\bm{\sigma}^2$, $p$, and $\pi_{y|\z}$ are parametrized by neural networks, learnable parameters being denoted by $\bm{\theta}$ and $\bm{\phi}$. Note that the classifier can handle binary, multi-class, and multi-label classification using a sigmoid, softmax, or multiple sigmoid activation function respectively at the output layer.

\begin{figure}
\centering
\includegraphics[scale=0.85]{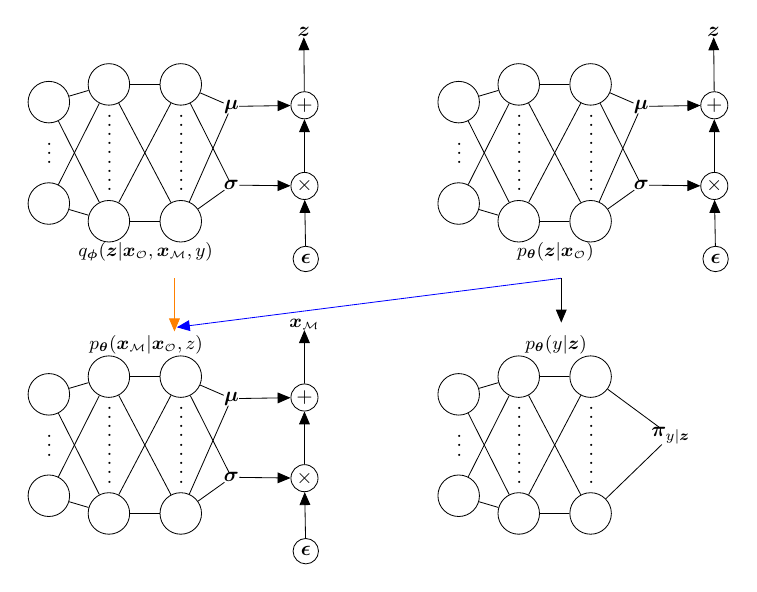} 
\vspace{-1.em}
\caption{Forward propagation in our proposed CMMD model. The orange arrow indicates a forward pass during training, which is replaced by the blue arrow at test time, i.e. the input to $p(\x_{\smallM}|\x_{\smallO},\z)$ is $\z \sim q(\z|\x_{\smallO},\x_{\smallM},y)$ during training, while  $\z \sim p(\z|\x_{\smallO})$ at test time. The black arrow depicts a common forward propagation during training and testing, i.e. the input to $p(y|\z)$ is always $\z \sim p(\z|\x_{\smallO})$.}
\vspace{-1em}
\label{fig_model}
\end{figure}

We observed that $\z \sim q(\z|\x_{\smallO},\x_{\smallM},y)$ leads to an unstable classification of $y$. We therefore fed the classifier $p(y|\z)$ with $\z \sim p(\z|\x_{\smallO})$ during training and test time. We hypothesize that the prior distribution reproduces the test scenario more accurately than the posterior distribution. Fig. \ref{fig_model} shows the forward propagation during training and test time in our proposed methodology\footnote{Full code of the model is available at \url{https://github.com/rogelioamancisidor/cmmd}}.

\section{Experiments and Results}
This section compares the CMMD model we propose with different multi-modal learning algorithms in downstream classification tasks across three different domains: image-to-image using a multi-modal version of MNIST and SVHN, acoustic-to-articulatory with the XRMB data set, and image-to-annotation using the MIR Flickr data set. The benchmark models are: CCA \citep{Hotelling1936}, DCCA \citep{andrew2013deep}, DCCAE \citep{wang2015deep}, MVCL \citep{hermann2013multilingual}, RBM-MDL \citep{ngiam2011multimodal}, VCCA \citep{wang2016deep}, MVAE \citep{wu2018multimodal}, and MMVAE \citep{shi2019variational}. We include a classifier model $\text{M-} \x_{\smallO}$ that only uses the always available modality, to allow the impact of joint representations for classification to be assessed. Network architecture and model training details are given in Appendix C. \review{However, given the importance of the $\omega$ hyperparameter in the optimization of our proposed model, we mention here the value found by cross-validation in each experiment, unless otherwise specified. See Figure \ref{app_all_omegas} for an overview over all $\omega$ values.}  

\subsection{Data sets}
In the following we explain the multi-modal data sets used in this research. 

\textbf{2-modality MNIST}: This data set, introduced by \cite{wang2015deep}, consists of 28 $\times$ 28 MNIST hand-written digit images \citep{deng2012mnist}. The images have been randomly rotated at angles in the interval $[-\pi / 4,\pi/4]$, to generate $\x_{\smallO}$. The modality $\x_{\smallM}$ is generated by randomly selecting a digit from $\x_{\smallO}$ and adding noise uniformly sampled from $[0,1]$ to each pixel in the non-rotated image. Each pixel is then truncated to the interval $[0,1]$.

\textbf{MNIST-SVHN}: We randomly paired each instance of a MNIST digit ($\x_{\smallO}$) with one instance of the same digit class in the SVHN data set \citep{netzer2011reading} ($\x_{\smallM}$), which is composed of street-view house numbers, just as in \cite{shi2019variational}.

\textbf{3-modality MNIST}: This data set combines some of the modalities in the previous data sets, i.e. original MNIST, rotated MNIST, and SVHN digits. All of the same digit class. 

\textbf{MNIST-SVHN-Text}: \review{This data set was first introduced in \cite{sutter2020multimodal} and it is based on the MNIST-SVHN data set. The additional string modality contains 8 characters where everything is a blank space except the digit word. Further, the starting position of the word is chosen randomly. The 8 character string is, finally, converted to a 71D one-hot-encoding, which corresponds to the length of possible characters in the dictionary used in \cite{sutter2020multimodal}. The experiments using this data set consider all possible combinations of missing and observable modalities, see Section \ref{sec_mst}.} 

\textbf{XRMB}: The original XRMB data set \citep{westbury1994x} contains simultaneously recorded speech and articulatory measurements from 47 American English speakers. The modality $\x_{\smallO}$, the acoustic data, is composed of a 13D vector of mel-frequency cepstral coefficients (MFCCs). We also included their first and second derivatives. This 39D vector is concatenated over a 7-frame window around each frame, resulting in a 273D vector that corresponds to $\x_{\smallO}$. The modality $\x_{\smallM}$, the articulatory data, is formed by horizontal and vertical displacements of 8 pellets on the tongue, lips, and jaw, resulting in a 112D vector. The data set then finally contains 40 phone classes.    

\textbf{Flickr}: The Flickr data set \citep{huiskes2008mir} contains 1 million images, 25000 being labeled according to 24 classes. Note that each image can be assigned to multiple classes. Stricter labeling was also carried out for 14 of the classes, images only being annotated with a category where that category was salient. The data set therefore has 38 classes. We used the same 3857D feature vector ($\x_{\smallO}$) as used by \cite{srivastava2012multimodal} to describe the images. The modality $\x_{\smallM}$ is composed of tags related to the image, the tags constrained to a vocabulary of the 2000 most frequent words. 

\subsection{Posterior Collapse in Multimodal Learning}\label{sec_vimodels}
This section evaluates the impact of posterior collapse in VCCA, MVAE, MMVAE and our proposed CMMD model. We therefore measured posterior collapse as the proportion of latent dimensions that are within $\textstyle \epsilon$ KL divergence of the prior for at least $99\%$ of the data sample, as introduced by \cite{lucas_dont_2019}.

We trained all models using a 4-fold cross-validation approach, each fold containing 2 speakers from the XRMB data set \citep{westbury1994x}. Table \ref{table_newexps} shows that CMMD, \review{optimized with $\omega=0.8$}, outperforms all other methods in terms of error rates and root mean square errors (rmse) for the generated missing modality. VCCA\footnote{In this experiment we learn the variance parameters for the decoder networks in VCCA for fair comparisons. The authors of the original paper that introduced VCCA, used fixed variance parameters. The results are given in Table \ref{table1}} surprisingly ranks number two in the classification task, despite having a simpler architecture than MVAE and MMVAE. MVAE has lower error rates than MMVAE, even when we train MMVAE using an importance weighted approach and $k = 10$ samples. MMVAE IWAE generates the missing modality more accurately than MMVAE ELBO, and achieves smaller error rates.

\begin{figure}[t!]
  \begin{minipage}[b]{\linewidth}
  \centering
  {\includegraphics[scale=.21]{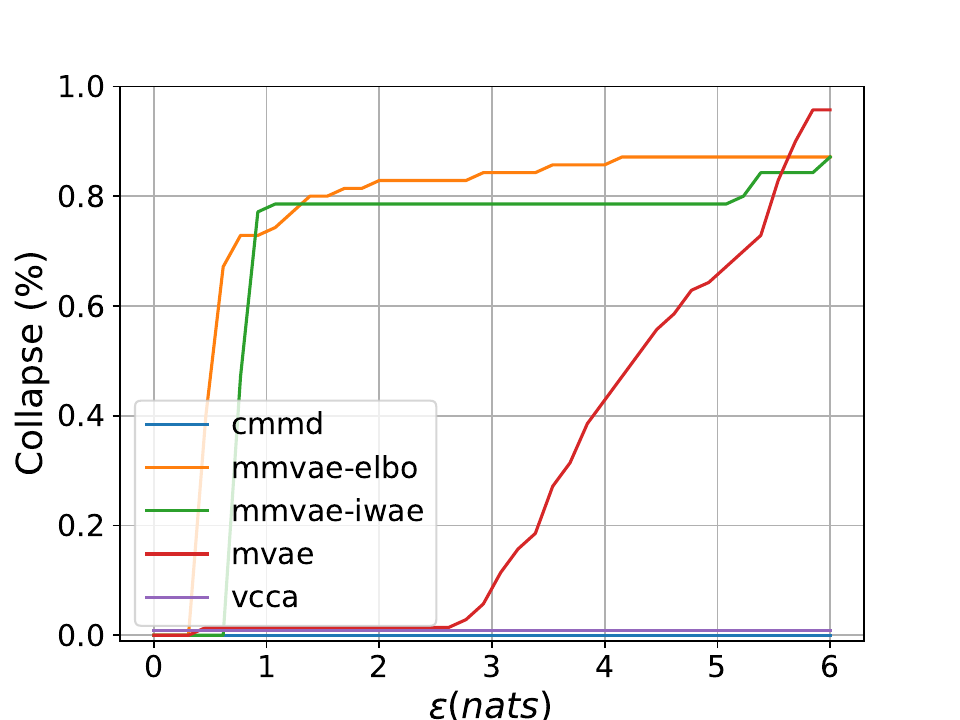}}
  {\includegraphics[scale=.21]{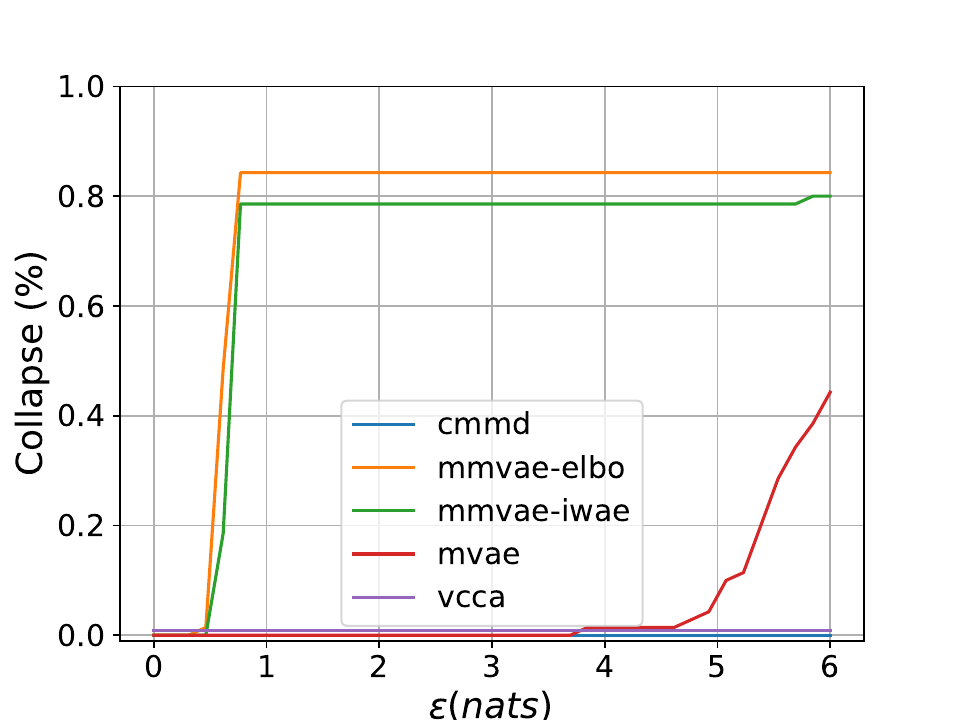}}
  {\includegraphics[scale=.21]{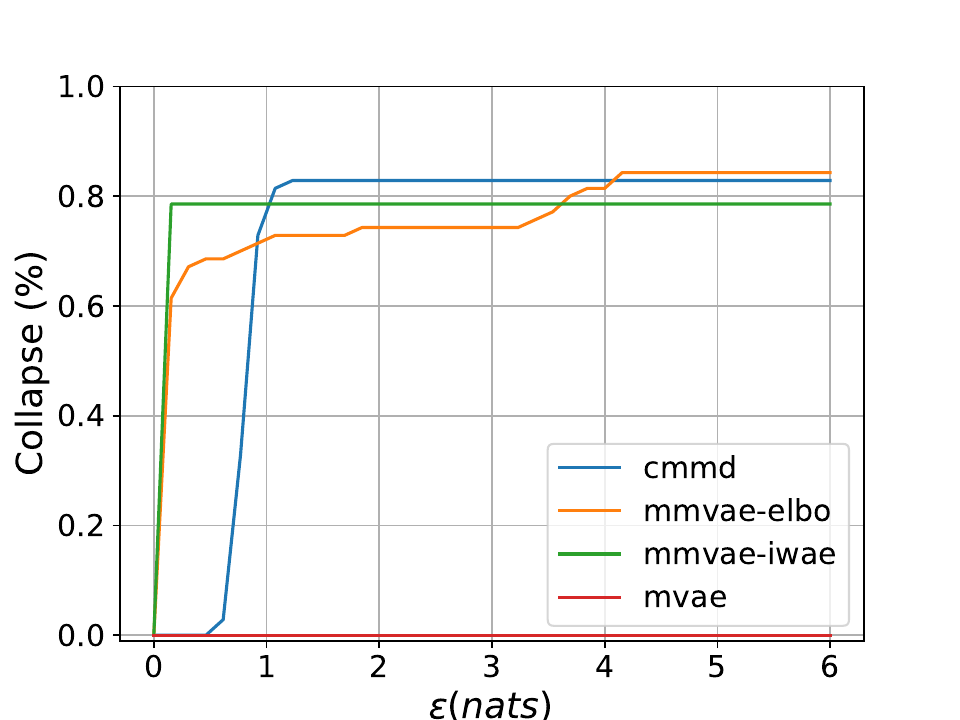}}
  {\includegraphics[scale=.21]{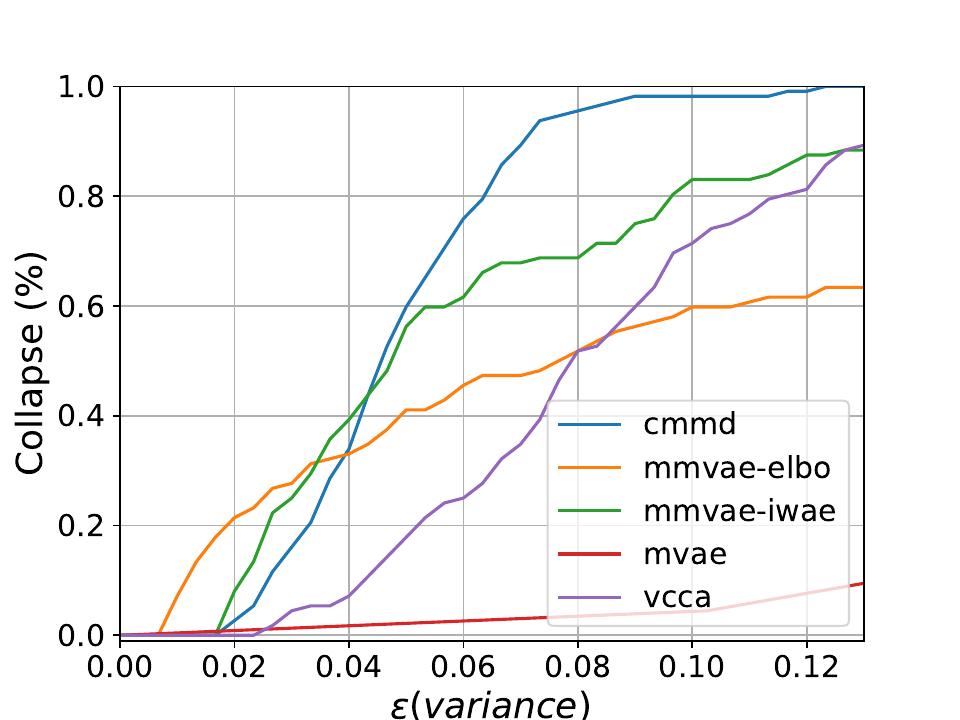}}
  \vspace{-0.5em}
  \caption{Posterior collapse (from left to right: $KL[(\z|\x_{\smallO})||\z]$, $KL[(\z|\x_{\smallM})||\z]$, and $KL[(\z|\x_{\smallO})||(\z|\x_{\smallM})]$) in VCCA, MVAE, MMVAE, and CMMD. The far right diagram shows the adoption of the concept of posterior collapse to measure the variance parameters in the decoder generating $\x_{\smallM}$. For example, at $\epsilon = 0.06$ around 79\% of the dimensions in $\hat{\x}_2$, generated by CMMD, have lower values than $\epsilon$.}
  \label{fig_collapse}
  \end{minipage}
  \vspace{-1em}
\end{figure}

The first two diagrams on the left side of Fig. \ref{fig_collapse} show the posterior collapse between $\z|\x_{\smallO}$ and $\z$, and between $\z|\x_{\smallM}$ and $\z$. They show, for both versions of MMVAE, that around $80\%$ of the dimensions in the latent representations collapse to $\mathcal{N}(\bm{0},\bm{1})$. This implies that the latent representation is independent of the observed modalities. MVAE, however, needs more than 5 nats when conditioned on the modality $\x_{\smallO}$, and more than 6 nats when conditioned on view $\x_{\smallM}$ before $80\%$ of the latent dimensions collapse. None of the latent dimensions in VCCA and CMMD are within 6 nats, and their latent representations therefore embed more information on the observed modalities. This information on the modalities is useful for downstream classification and, for CMMD, for the generation of the missing modality. The third diagram finally shows posterior collapse between the representations generated using $\z|\x_{\smallO}$ and $\z|\x_{\smallM}$. We want, in this case, $\z|\x_{\smallO}$ to collapse into $\z|\x_{\smallM}$, this meaning that the model is able to learn joint representations that contain information on $\x_{\smallM}$. Note that, in MMVAE, the collapse between both marginal distributions is strong given that both collapsed to $\mathcal{N}(\bm{0},\bm{1})$. On the other hand, the marginal distributions in MVAE embed information on the modalities (see first two diagrams). MVAE, however, fails to learn joint representation as suggested by the third diagram. CMMD does, however, counteract posterior collapse through the conditional prior and through directly optimizing mutual information, as shown by the first three diagrams. 

\begin{table}[t]
\centering
\caption{\textcolor{blue}{Error rates (\%)} and \textcolor{orange}{rmse} (lower is best) for the experiment using 4 randomly chosen folds (speakers IDs $[(1,3),(43,45),(10,13),(27,29)]$) from the XRMB data set, where the shared latent representation is generated using the available modality at test time. Note that VCCA cannot generate missing modalities in the scenario considered in this experiment. We add a baseline classifier M-$\x_{\smallO}$ that only uses $\x_{\smallO}$.}
\def\arraystretch{0.9}
\setlength{\tabcolsep}{9pt}
\begin{tabular}{|c|c|c|c|c|c|c|}
\hline
Fold & M-$\x_{\smallO}$ & VCCA  & MVAE  &   MMVAE ELBO & MMVAE IWAE & CMMD \\
\hline
   \multirow{2}{*}{1}   & \textcolor{blue}{$39.9$} & \textcolor{blue}{$40.2$} & \textcolor{blue}{$45.5$} & \textcolor{blue}{$54.9$} & \textcolor{blue}{$48.9$} & \textcolor{blue}{$\BB{32.4}$} \\
                        & -     &   -      & \textcolor{orange}{$1.07$} & \textcolor{orange}{$1.29$} & \textcolor{orange}{$0.77$} & \textcolor{orange}{$\BB{0.74}$} \\
                     \hline
    \multirow{2}{*}{2}  & \textcolor{blue}{$36.5$} & \textcolor{blue}{$40.6$} & \textcolor{blue}{$44.2$} & \textcolor{blue}{$49.4$} & \textcolor{blue}{$47.0$} & \textcolor{blue}{$\BB{31.2}$} \\
                        & -     &   -      & \textcolor{orange}{$1.06$} & \textcolor{orange}{$1.28$} & \textcolor{orange}{$0.79$} & \textcolor{orange}{$\BB{0.75}$} \\
                     \hline
    \multirow{2}{*}{3}  & \textcolor{blue}{$54.9$} & \textcolor{blue}{$55.4$} & \textcolor{blue}{$56.8$} & \textcolor{blue}{$61.9$} & \textcolor{blue}{$60.2$} & \textcolor{blue}{$\BB{47.4}$} \\
                        & -     &   -      & \textcolor{orange}{$1.09$} & \textcolor{orange}{$1.17$} & \textcolor{orange}{$0.83$} & \textcolor{orange}{$\BB{0.77}$} \\
                     \hline
    \multirow{2}{*}{4}  & \textcolor{blue}{$48.0$} & \textcolor{blue}{$47.2$} & \textcolor{blue}{$51.7$} & \textcolor{blue}{$59.1$} & \textcolor{blue}{$53.7$} & \textcolor{blue}{$\BB{38.2}$} \\
                        & -     &   -      & \textcolor{orange}{$1.07$} & \textcolor{orange}{$1.23$} & \textcolor{orange}{$0.82$} & \textcolor{orange}{$\BB{0.80}$} \\
\hline
  \multirow{2}{*}{Avg.}  & \textcolor{blue}{$44.8$} & \textcolor{blue}{$45.8$} & \textcolor{blue}{$49.6$} & \textcolor{blue}{$56.3$} & \textcolor{blue}{$52.5$} & \textcolor{blue}{$\BB{37.3}$} \\
                         & -     &   -      & \textcolor{orange}{$1.07$} & \textcolor{orange}{$1.24$} & \textcolor{orange}{$0.80$} & \textcolor{orange}{$\BB{0.76}$} \\
\hline
\end{tabular}
\label{table_newexps}
\vspace{-1.2em}
\end{table}
We adapted the posterior collapse definition to the analysis of the variance parameters in the generative process, to allow us to understand the rmse results for the generated missing modality. This insight is shown in the last diagram of Fig. \ref{fig_collapse}. For example at $\epsilon = 0.06$, around $79\%$ of the dimensions in $\hat{\x}_2$ generated by CMMD, have lower values than $\epsilon$. Furthermore, only $45\% $ of the parameters learned by MMVAE ELBO have lower values than $\epsilon$. We therefore hypothesize that MMVAE ELBO and MVAE overestimate the variance parameters in the decoder, resulting in higher rmse. The significant improvement for MMVAE IWAE seems to only change the decoder to a high capacity decoder and does not really improve the learned representations. Note that the variance collapse for VCCA is included for reference. It is actually generated using the modality $\x_{\smallM}$, which in theory is missing. 

A mixture of experts and product of experts provide an elegant cross-generation in multi-modal learning, the joint posterior distribution being a linear combination of marginal parameters or distributions. Our approach to learning the posterior distribution is, however, to use a single encoder network, which can capture interactions between all modalities. The model we propose handles missing modalities using a conditional prior modulated by the available modalities. VCCA presents an interesting alternative to learning joint representations, the generative process embedding information on modalities into $\z$. VCCA cannot, however, generate missing modalities, which its generative model requires. Note that only CMMD has lower error rates than the baseline model M-$\x_{\smallO}$, which indicates that current variational multi-modal models are not suitable for learning useful joint representations for downstream classification. CMMD should therefore be preferred over VCCA, MVAE and MMVAE given that, in the setting of this research, CMMD outperforms concurrent models in downstream classification and in the generation of missing modalities at test time.
\begin{table}[t]
\centering
\caption{We report error rates (lower is best) for experiments with MNIST and XRMB (average over speakers in the test dataset). For the Flickr data set, we report the mean average precision (mAP; higher is best). Results are based on \cite{wang2016deep}, except for values marked with $\dagger$ (which are from our own tests without pre-trained weights with Boltzmann machines) and results for CMMD.}
\def\arraystretch{0.95}
\setlength{\tabcolsep}{10pt}
\begin{tabular}{|c|c|c|c|c|}
\hline
Model name & Pretrain & MNIST error($\%$) & XRMB error($\%$) & Flickr mAP($\%$)\\
\hline
M-$\x_{\smallO}$          & \ding{55} & $13.1$  & $37.6$ & $48.0$ \\
\hline
DCCA        & \ding{51} & $2.9$   &  - & - \\
DCCAE       & \ding{51} & $2.2$   &  -  & - \\
\hline
CCA         & \ding{55} & $19.1$  & $29.4$ & $52.9$ \\
DCCA   & \ding{55} & $4.7^{\dagger}$     &  $25.4$        & $57.3$        \\
DCCAE   & \ding{55} & $4.4^{\dagger}$     &  $25.4$        &  $57.3$        \\
MVCL        & \ding{55} & $2.7$   & $24.6$ & $56.5$ \\
RBM-MDL     & \ding{55} & $11.7$  & $29.4$ & $47.7$ \\
VCCA        & \ding{55} & $3.0$   & $28.0$ & $60.5$ \\
VCCA-private & \ding{55} & $\BB{2.4}$   & $25.2$ & $61.5$ \\
MVAE        & \ding{55} & $6.0$     & $39.8$   & $\BB{65.0}$ \\
MMVAE IWAE  & \ding{55} & $12.3$     & $37.4$   & $50.0$ \\
CMMD        & \ding{55} & $\BB{2.4}$     & $\BB{21.1}$   & 64.1 \\
\hline
\end{tabular}
\label{table1}
\vspace{-1.em}
\end{table}

\subsection{Classification and Generating the Missing Modality}
\subsubsection{Image-to-Image with MNIST}\label{sec_mm}
Table \ref{table1} shows that the performance of our proposed CMMD model is on a par with state-of-the-art models, including those that use pre-trained weights. We observed (practically) the same model performance for this data set at different $\omega$ values, our best model using a value of 0.4. Note that both DCCA and DCCAE use pre-trained weights with Boltzmann machines (BMs) \citep{salakhutdinov2009deep}. We therefore, for completeness, also retrained DCCA and DCCAE without using pre-trained weights. 2D t-SNEs of the latent space can be found in Appendix F.

We used (in a second analysis) the original version of MNIST as $\x_{\smallO}$ and the SVHN data set as $\x_{\smallM}$. Our best model used $\omega=0.1$ and achieved a higher accuracy than MVAE and MMVAE, as shown in Table \ref{tbl_mnist_shvn}.

To show that CMMD can handle more than one missing and observed modality, we construct a 3-modality data set matching the class labels in: MNIST ($\x_1$), rotated MNIST ($\x_2$), and SVHN ($\x_3$). We used the same model parameters as were used in the previous experiment, and considered two test scenarios: i) rotated MNIST and SVHN are both  missing, i.e. $\x_{\smallO}=\x_1$ and $\x_{\smallM}=(\x_2,\x_3)$ and ii) SVHN is missing, i.e. $\x_{\smallO}=(\x_1,\x_2)$ and $\x_{\smallM}=\x_3$. The top (bottom) row in Table \ref{tbl_3mexps} shows the classification performance for the test scenario, in which two (one) modalities are missing. Generated modalities are shown in Appendix G. 

\subsubsection{Image-to-Text with MNIST and SVHN}\label{sec_mst}
\review{Note that given $M=3$ modalities, there are $2^M-1=7$ combinations of observable modalities $\x_{\smallO}$\footnote{\review{These combinations are \{M\},\{S\},\{T\},\{M,S\},\{M,T\},\{S,T\}, and \{M,S,T\}, where M, S, and T refer to the MSNIT, SVHN, and Text modalities, respectively.}}. We generate multimodal representations, conditioned on all of the possible combinations of observed modalities, with the CMMD model. After training, we randomly choose 500 representations from the training data set to train a multiclass logistic regression to classify true digits. Table \ref{tbl_allelbos} compares the classification performance of the CMMD model (see Figure \ref{app_all_omegas} and Table \ref{tbl_omega_cross} to know the $\omega$ values used in these experiments), under this two-step classification approach, with that of MVAE, MMVAE, and MoPoE in similar experiments to those in \cite{sutter2021generalized} and \cite{javaloy2022mitigating}. We report model accuracy averaged over all 7 combinations of observable modalities and 5 different runs. Models ending with \textit{IO} are trained with the impartial optimization approach introduced and reported in \cite{javaloy2022mitigating}.}

\begin{figure}[t!]
    \centering
    \subcaptionbox*{{(a) MNIST and SVHN are missing}}{\includegraphics[scale=.53]{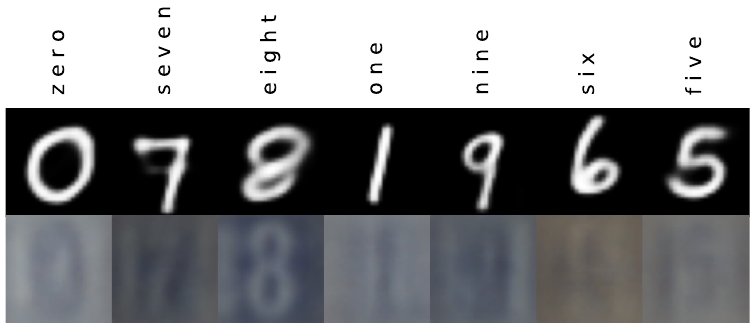}}
    \centering
    \subcaptionbox*{(b) SVHN is missing}{\includegraphics[scale=.53]{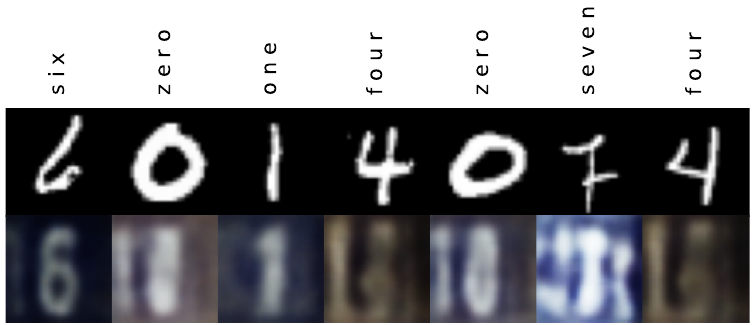}}
    \caption{\review{The left panel shows generated images for the MNIST and SVHN modalities conditioned on the observed Text modality using the CMMD model. The right panel shows generated images for the SVHN modality conditioned on Text and MNIST modalities, which are assumed to be observed at test time.}}
    \label{fig_missingmnist_svhn}
\end{figure}

\review{We can see that IO increases the classification accuracy of all three models, especially for MMVAE. However, CMMD achieves higher discriminative power in all scenarios of observable modalities. Furthermore, Figure \ref{fig_missingmnist_svhn} shows some examples of the images generated by CMMD. The left panel shows generated images for MNIST and SVHN modalities conditioned on the Text modality, while the right panel shows generated images for the SVHN modality conditioned on both Text and MNIST modalities. Note that the SVHN images in the right panel are sharper, compared to the left panel, since the generative model is conditioned on more observed modalities in that case. Details on model architectures and hyperparameter values are in Appendix H.}

\review{Finally, using the same models as before, we evaluate the quality of the generated missing modalities conditioned on all different combinations of observable modalities, i.e. conditional or cross-modal generation. To that end, we use the generative coherence metric, first introduced in \cite{shi2019variational}. Following previous works and same architectures as in \cite{sutter2021generalized}, we train a classifier on the original unimodal training data set to classify the generated modalities. If the classifier detects the same attributes in the generated samples, it is a coherent generation. Further, we use classification accuracy to measure the quality of generated samples. Table \ref{tbl_coherence} shows accuracy values of the conditionally generated modalities averaged over 5 different runs. The letter at the top indicates the modality being generated based on the
different sets of modalities below, where M, S, and T stands for MNIST, SVHN, and Text modalities. CMMD achieves higher accuracy in most of the conditional generation scenarios.}

\begin{table}
\centering
\begin{minipage}[t]{0.48\linewidth}\centering
\caption{Accuracy results for downstream classification with MNIST-SVHN. Results for MVAE and MMVAE are based on \cite{shi2019variational}.}
\begin{tabular}{|c|c|}
\hline
Model &  MNIST-SVHN \\
name &   accuracy (\%) \\    
\hline
 MVAE        & $95.7$ \\
 MMVAE       & $91.3$ \\
 CMMD        & $\BB{97.6} \pm 0.08\%$ \\
\hline 
\end{tabular}
\label{tbl_mnist_shvn}
\end{minipage}\hfill%
\begin{minipage}[t]{0.48\linewidth}\centering
\centering
\caption{Accuracy results for 3-modality MNIST. The first experiment classifies using representations generated with $\x_{\smallO}=\x_1$, while the second experiment uses $\x_{\smallO}=(\x_1,\x_2)$}
\begin{tabular}{|c|c|}
\hline
Missing &  Accuracy (\%) \\
Modality &   \\    
\hline
$\x_{\smallM}=(\x_2,\x_3)$  & $97.5 \pm 0.25\%$ \\
$\x_{\smallM}=\x_3$         & $98.9 \pm 0.13\%$ \\
\hline 
\end{tabular}
\label{tbl_3mexps}
\end{minipage}
\end{table}

\begin{table}[t!]
\caption{\review{Accuracy performance (\%), averaged over all 7 combinations of observable modalities and 5 different runs with the MNIST-SVHN-Text data set, of the MVAE, MMVAE, MoPoE, and CMMD models. Additionally, we include the results from \cite{javaloy2022mitigating} where MVAE, MMVAE, and MoPoE are trained with impartial optimization.}}
\centering
\def\arraystretch{0.95}
\setlength{\tabcolsep}{10pt}
\begin{tabular}{|c|c|c|c|}
\hline
Model & \cite{javaloy2022mitigating} & \cite{sutter2021generalized} & Ours \\
\hline 
MVAE        & 69.7  & 83.1   & -    \\
MVAE-IO     & 70.0  & -      & -    \\
MMVAE       & 87.6  & 89.0   & -    \\
MMVAE-IO    & 90.8  & -      &  -   \\
MoPoE       & 89.9  & 95.1   &  -   \\
MoPoE-IO    & 91.5  &   -    &  -   \\
CMMD        & -     &  -     & \BB{96.5} \\
\hline
\end{tabular}
\label{tbl_allelbos}
\end{table}

\begin{table}[t!]
\caption{\review{Accuracy values of the conditionally generated modalities averaged over 5 different runs. The letter at the top indicates the modality being generated based on the different sets of modalities below, where M, S, and T stands for MNIST, SVHN, and Text modalities, respectively.}}
\centering
\def\arraystretch{0.95}
\setlength{\tabcolsep}{10pt}
\begin{tabular}{|c|ccc|ccc|ccc|}
\hline
    & \multicolumn{3}{c|}{M} & \multicolumn{3}{c|}{S} & \multicolumn{3}{c|}{T} \\
\cline{2-10}    
Model & S & T & S,T & M & T & M,T & M & S &  M,S \\
\hline 
MVAE        & 0.24 & 0.20 & 0.32 & 0.43 & 0.30 & 0.75 & 0.28 & 0.17 & 0.29  \\
MVAE-IO     & 0.11 & 0.26 & 0.28 & 0.50 & 0.33 & 0.30 & 0.61 & 0.12 & 0.64\\
MMVAE       & \BB{0.75} & 0.99 & 0.87 & 0.31 & 0.30 & 0.30 & 0.96 & \BB{0.76} & 0.84 \\
MMVAE-IO    & 0.49 & 0.79 & 0.64 & \BB{0.87} & 0.76 & 0.82 & 0.97 & 0.58 & 0.77 \\
MoPoE       & 0.74 & 0.99 & 0.94 & 0.36 & 0.34 & 0.37 & 0.96 & \BB{0.76} & 0.93 \\
MoPoE-IO    & 0.11 & 0.63 & 0.52 & 0.28 & 0.47 & 0.43 & 0.80 & 0.11 & 0.90 \\
CMMD        & \BB{0.75} & \BB{1.00} & \BB{1.00} & 0.66 & \BB{0.87} & \BB{0.87} & \BB{0.98} & 0.69 & \BB{0.98} \\
\hline
\end{tabular}
\label{tbl_coherence}
\end{table}

\subsubsection{Acoustic-to-Articulatory with XRMB}\label{xrmb_exps}
The experimental setup and the data pre-processing used in this section are based on \cite{wang2016deep}. Table \ref{table1} shows average error rates for all test speakers, CMMD outperforming previous models without a domain-specific classifier. \review{For this experiment, CMMD is optimized with $\omega=0.7$}. Note that \cite{wang2016deep} used the tandem speech recognizer \citep{hermansky2000tandem} as classifier model in all the experiments they conducted. The tandem speech recognizer successfully couples neural networks and Gaussian mixtures models for word recognition and, in the benchmark results of \cite{hermansky2000tandem}, reduced speech classification error rates by 35\%. \cite{wang2016deep} also used the 39D vector of MFCCs and the joint data representations as input data for the tandem recognizer for all experiments. We hypothesize that this further improves the performance of the tandem recognizer. The CMMD model we propose, however, only uses the shared data representations for classification\footnote{In the current version of the data set, it is not possible to identify the 39D vector of MFCCs in the 273D vector for the modality $\x_{\smallO}$. We could therefore not concatenate MFCCs and joint representations for training the classifier.}.

\subsubsection{Image-to-Annotation with Flickr}
We use the same data set in this section as that used in \cite{srivastava2012multimodal}. Most of the Flickr data corresponds to unlabeled images. We therefore used a two-stage training approach. Firstly, we trained our proposed model, but without the classifier and omitting the class label in the encoder, i.e. $q(\z|\x_{\smallO},\x_{\smallM})$. Secondly, we used the weights from the first stage in the corresponding networks of Eq. \ref{eq_convex_elbo} and used random weights at initialization for $y$ in the encoder $q(\z|\x_{\smallO},\x_{\smallM},y)$.

Following the standards set by previous research, we use the mean average precision (mAP) to measure the classification performance of our proposed CMMD model for 10000 randomly selected images. Table \ref{table1} shows that CMMD, \review{optimized with $\omega=0.5$}, and MVAE outperform previously proposed image classification methods. 
\begin{table*}[t]
    \centering
    \caption{Tags describing images are generated with the multi-modal learning deep Boltzmann machine (DBM)  \citep{srivastava2012multimodal} and with CMMD. DBM fails to generate coherent tags in the first 3 images. CMMD is, however, able to generate meaningful tags. In the last image, both models generate coherent tags.}
    \small
    \begin{tabular}{c|ccccc}
        & \includegraphics[height=1.6cm,keepaspectratio]{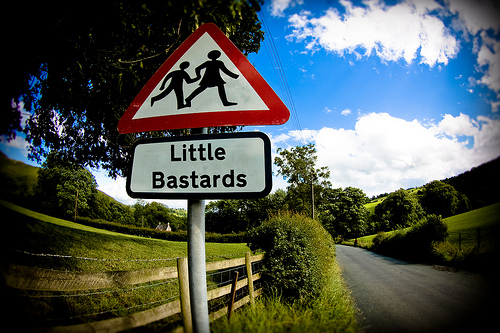} 
        & \includegraphics[height=1.6cm,keepaspectratio]{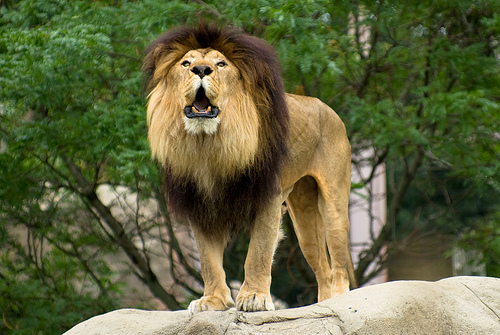} 
        & \includegraphics[height=1.6cm,keepaspectratio]{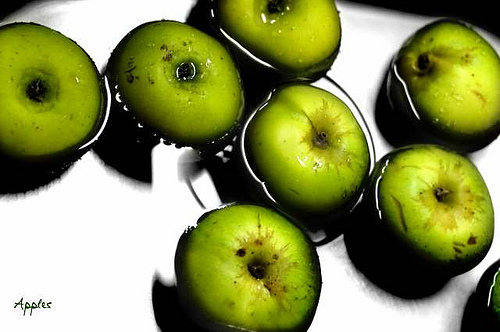} 
        & \includegraphics[height=1.6cm,keepaspectratio]{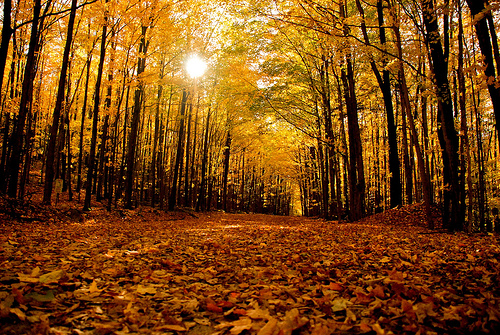} 
                        \\
        \hline
        Generated tags  & water, glass, wine, 
                        & portrait, women,
                        & nikon, d200,   
                        & foliage, autumn,
                        \\
        DBM             & drink, beer, 
                        & soldier, postcard 
                        & tamron, d300, 
                        & trees, leaves,
                        \\
                        & bubbles, splash, 
                        & soldiers, army
                        & f28, sb600, d60 
                        & fall, forest,
                        \\
                        & drops, drop 
                        & 
                        & nikkor, d50, d90 
                        & woods, path
                        \\
        \hline
        Generated tags  & -
                        & statue
                        & car, performance
                        & a700
                        \\
        MVAE            & 
                        & 
                        & 
                        & 
                        \\
        \hline
        Generated tags  & canon, night, 2007
                        & nikon, green, lion
                        & flower
                        & trees, autumn
                        \\
        MMVAE IWAE      & 
                        & 
                        & 
                        & 
                        \\
        \hline
        Generated tags  & sign, fisheye 
                        & animal, lion,  
                        & apple, food
                        & nature, light
                        \\
        CMMD            & 
                        & outdoors, zoo, k10d,   
                        & 
                        & autumn, leaves
                        \\
                        & 
                        & challengeyouwinner,  
                        & 
                        & wood, path,
                        \\
                        & 
                        & boston, wildlife
                        &
                        & forest
                        \\
    \end{tabular}
    \label{table3}
\vspace{-2em}
\end{table*}
\subsubsection{Acoustic Inversion and Annotation Generation}\label{sec_ai_ag}
We tested the generative process $p(\x_{\smallM}|\x_{\smallO},\z)$ in CMMD in image-to-annotation mapping  and also acoustic-to-articulatory (called acoustic inversion (AI)). The scarce availability of articulatory data \citep{badino2017speaker} makes acoustic inversion an important field. Table \ref{table2} shows, on the test data set, that CMMD outperforms the rmse for AI reported in \cite{wang2015unsupervised}, which is based on the training and validation data set ($1.17$ and $1.96$, respectively). Our results also outperform the average rmse of $2.14$ obtained on the test dataset of \cite{badino2017speaker}.

The second experiment involves generating tags, which can be costly to obtain, that describe a given picture in the Flickr data set. We used our trained model from the previous section and compared it with the deep Boltzmann machine (DBM) model \citep{srivastava2012multimodal}, MVAE, and MMVAE. We furthermore tested all models on different images and with different levels of complexity. Table \ref{table3} shows some of the generated tags. More examples are given in Appendix D. 

The generative process in CMMD generates quality articulatory and annotation samples at test time. The results suggest that the prior distribution in our proposed model learns joint representations through the optimization of our proposed objective function, which maximizes mutual information between representations and the missing modality at test time.

\subsection{Analysis of the Objective Function}\label{sec_ablation}
In this section, we train the CMMD model using the XRBM data set for all speakers in Table \ref{table2} using a speaker-dependent approach, i.e. $70\%-30\%$ of the data for each speaker used for training-testing, \review{unless otherwise specified}. We furthermore trained the CMMD model in two ways: i) we fine-tuned $\omega$ in the range $[0, 0.1, \cdots, 1]$ and ii) we used $\omega = 0.7$ (which is the optimal value in the previous section) and fixed the variance parameters in the decoder network $p(\x_{\smallM}|\x_{\smallO},\z)$ to the same value as in \cite{wang2016deep}, i.e. $\sigma^2=0.01$.

\vspace{-0.5em}
\paragraph{Impact of Fixed Variance Parameters:} 

\begin{table}[t]
\centering
\caption{We report rmse for AI and error rates (\%) for downstream classification in a speaker-independent experiment for eight speakers. Average and standard deviation (std) values are shown at the bottom.}
\def\arraystretch{0.95}
\setlength{\tabcolsep}{7pt}
\begin{tabular}{|cccccc|}
\hline
\multicolumn{6}{|c|}{XRMB}\\
\hline
Speaker     & AI    & Classification & Speaker     & AI    & Classification\\
ID          & rmse  & error (\%)& ID          & rmse  & error (\%)\\
\hline
$7$&$0.80$&$15.8$&$23$&$0.79$&$26.8$\\
$16$&$0.76$&$21.4$&$28$&$0.57$&$22.5$\\
$20$&$0.73$&$16.2$&$31$&$0.77$&$20.6$\\
$21$&$0.78$&$25.8$&$35$&$0.78$&$20.2$\\
\hline
\footnotesize average 8 speakers & $0.75$ & $21.2$& \footnotesize std 8 speakers & $0.07$ & $3.7$ \\
\hline
\end{tabular}
\label{table2}
\end{table}

\begin{figure}[t!]
    \centering
    \includegraphics[scale=0.4]{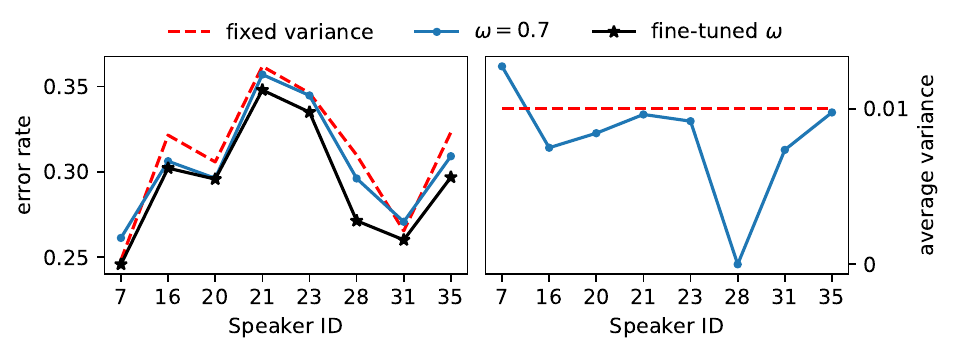}
    \includegraphics[scale=0.4]{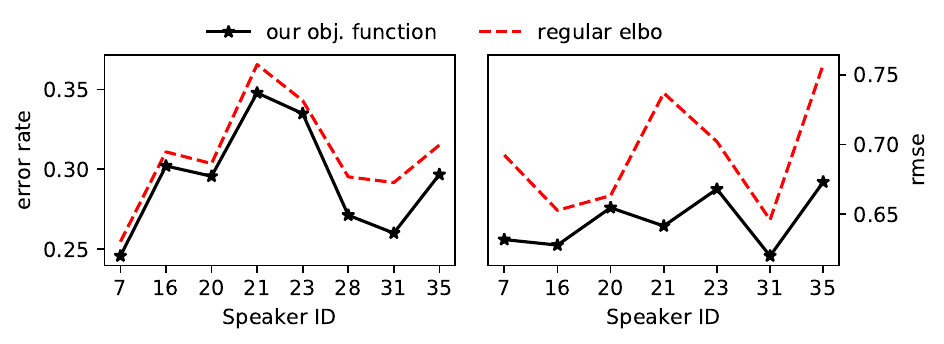}
    \vspace{-0.5em}
    \caption{
    The 1st and 3rd plot show error rates for the speaker-dependent experiments (Section \ref{sec_ablation}). The 2nd plot shows average variances of all generated $\x_{\smallM}$ features. The last plot compares rmse for their generated values. Speaker 28 was removed as the rmse in both cases is roughly 0.
    }
    \label{fig_fixedvar}
    \vspace{-1.em}
\end{figure}

The second diagram in Fig. \ref{fig_fixedvar} shows some between-variability in the modality $\x_{\smallM}$. Fixing the variance parameters in $p(\x_{\smallM}|\x_{\smallO},\z)$ therefore deteriorates error rates, as shown in the first diagram. 

\begin{figure}[t!]
  \begin{minipage}[b]{\linewidth}
  \centering
  {\includegraphics[scale=.3]{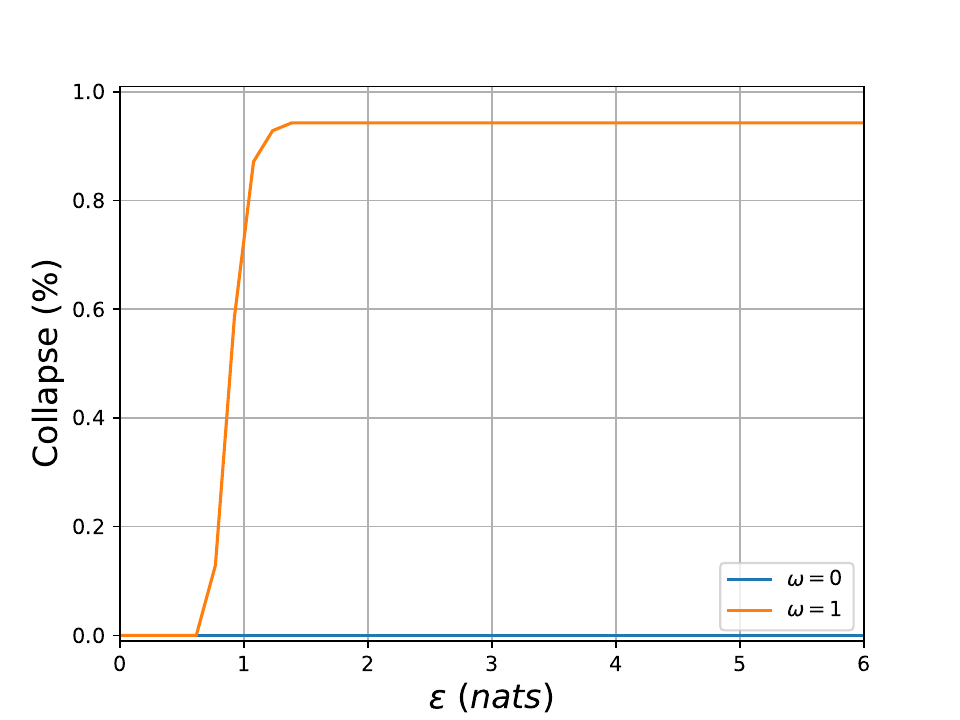}}\hspace{27em}
  {\includegraphics[scale=.27]{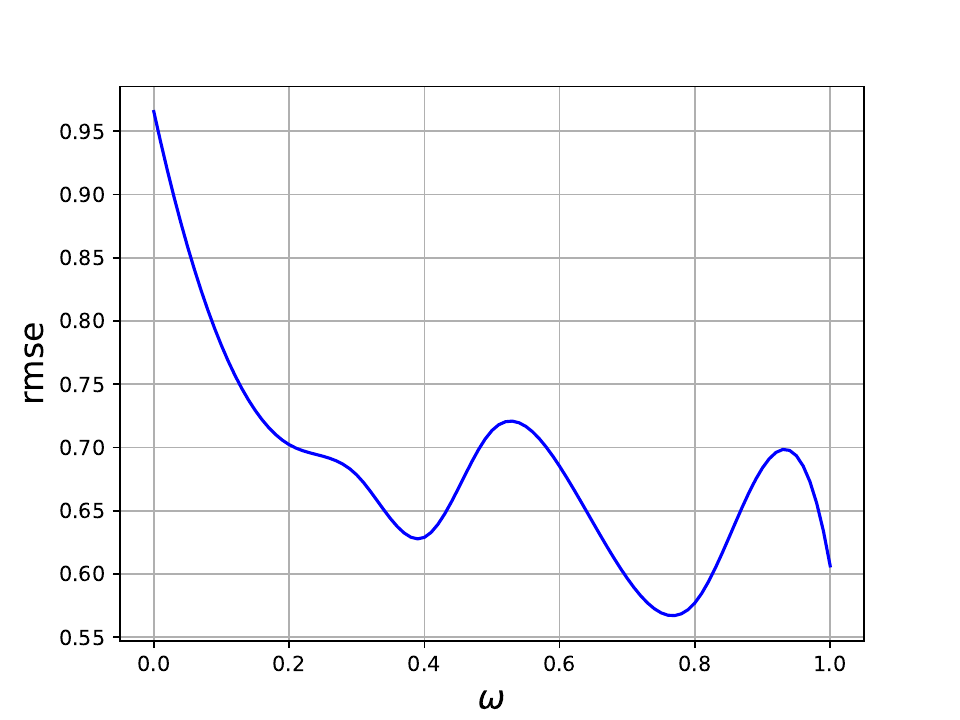}}
  {\includegraphics[scale=.27]{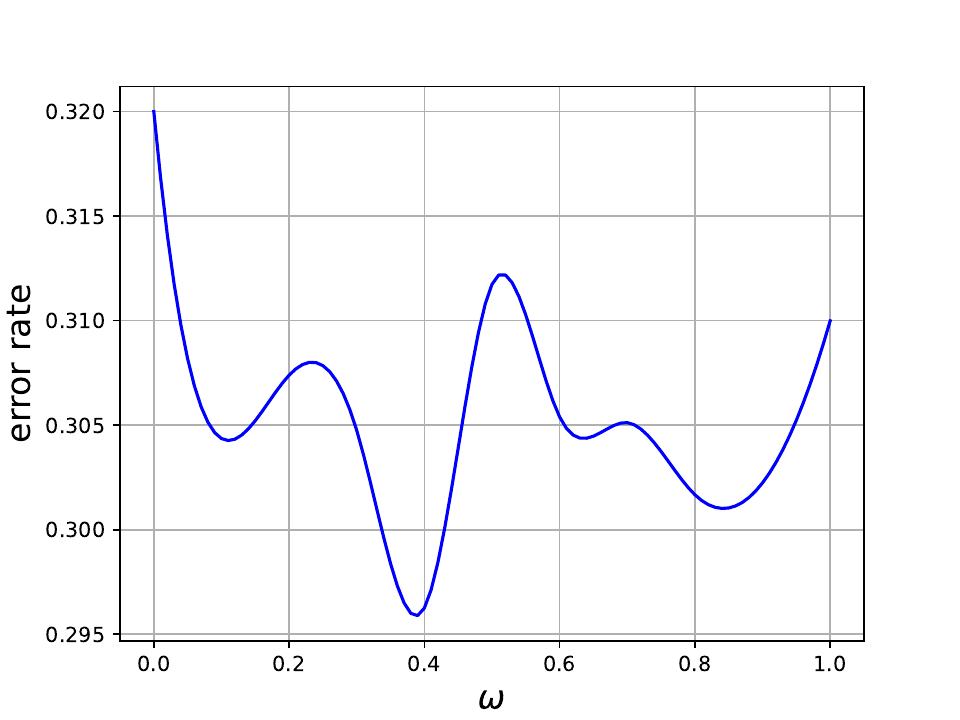}}\hspace{27em}
  {\includegraphics[scale=.27]{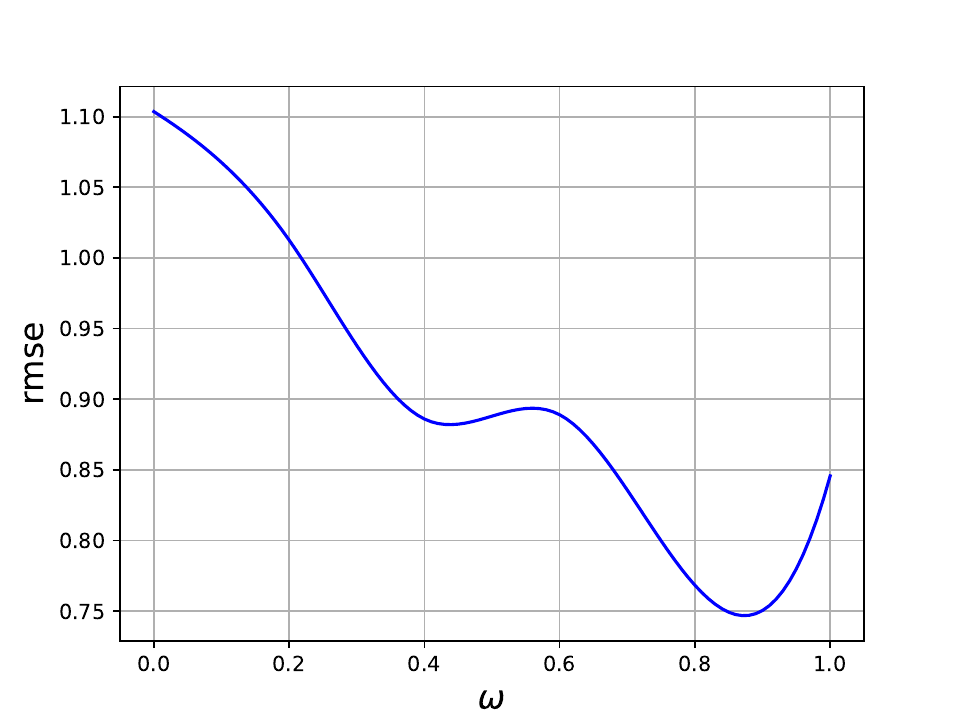}}
  {\includegraphics[scale=.27]{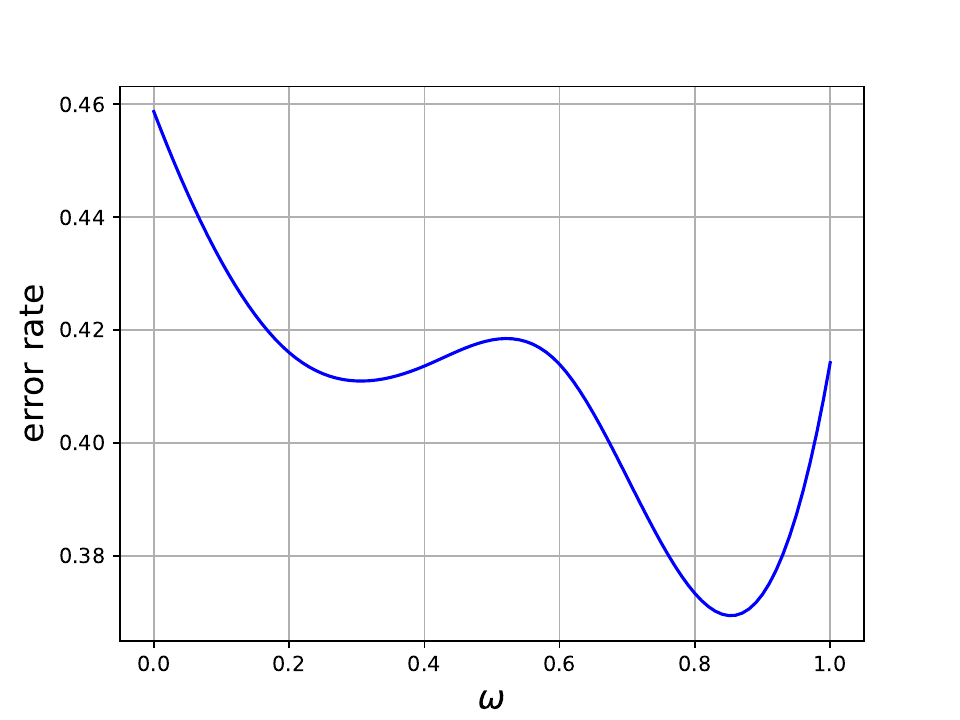}}
  \vspace{-0.5em}
  \caption{\review{The top panel shows the posterior collapse in the CMMD model for $\omega=0$ and $\omega=1$. In both cases, we use data for speaker 7 in the XRMB data set. The two panels in the middle show average rmse and error rate as a function of $\omega$ for speakers 7, 16, 20, 21, 23, 28, 31, and 35 in the XRMB data set. Finally, the two panels in the bottom show average rmse and error rate values in the cross-validation approach introduced in Section \ref{sec_vimodels}.}}
  \label{fig_collapse_nn}
  \end{minipage}
  \vspace{-2.5em}
\end{figure}

\vspace{-1em}
\paragraph{Should we optimize the ELBO?}
The third panel in Fig. \ref{fig_fixedvar} compares error rates for the ELBO (dashed line), recovered for $\omega=1$, and for our proposed objective function (Eq. \ref{eq_convex_elbo}) with fine-tuned $\omega$. Our proposed objective function achieves lower error rates for all speakers. The rmse for the generated features in $\x_{\smallM}$ are also smaller when we optimize our proposed objective function. 

\review{The top panel in Figure \ref{fig_collapse_nn} shows the posterior collapse in the CMMD model for $\omega=0$ and $\omega=1$; the latter optimizes the ELBO, while the former optimizes mutual information (in addition to the generative and classifier models). Remember that the main motivation for including the mutual information term $I(\x_{\smallM},\z|\x_{\smallO})$ is to counteract the posterior collapse problem and, from the figure, it is clear that CMMD avoids the posterior collapse problem by optimizing mutual information. However, as shown by the four panels in the middle and bottom of Figure \ref{fig_collapse_nn} (in which we add the relatively more complex learning task presented in Section \ref{sec_vimodels}, but varying $\omega$) optimizing only mutual information harms the performance of the generative and classifier models, reflected in the rmse and error rate respectively. Note that only optimizing mutual information accounts for the minimization of an average MMD divergence measure. That is, we only minimize the divergence from the average conditional posterior $q(\z|\x_{\smallO})$ to the conditional prior. Our results confirm that minimizing an average divergence measure makes the prior distribution, which is used for downstream tasks, unable to sculpt latent representations as suggested by \cite{hoffman2016elbo}. On the other hand, only optimizing the term-by-term KL divergence leads to latent representations $\z$ that are independent from $\x_{\smallM}$, which turns out to be relatively less harmful for downstream tasks. Fortunately, our proposed objective function offers a way of trading-off these two effects and, as can be seen in the middle and bottom rows in Figure \ref{fig_collapse_nn}, there is an $\omega$ region in which the generative and classifier models achieve higher performance. Hence, the optimal $\omega$ value is specific to the learning task and must be found by cross-validation.} 

\vspace{-1em}
\paragraph{How much overhead does mutual information optimization add?} \review{We use the MNIST-SVHN-Text data set to measure training time for $\omega=0$ and $\omega=1$. The average training time for processing one batch with 256 observations is 10.59 milliseconds if the ELBO is optimized. On the other hand, the average training time to optimize our proposed objective function, including mutual information, is 11.04 milliseconds, which is the same training time for $1>\omega>0$. Therefore, our proposed objective function does not add significant overhead and is able to achieve higher performance in the downstream tasks considered in this research.}

\section{Conclusion}
This research studies the effect of posterior collapse in downstream classification and in the generative process of multi-modal learning models. We show that the variational lower bound on the conditional likelihood has a Kullback-Leibler divergence that limits the amount of information on the modalities embedded in the joint representation. We, to counteract this effect, propose a novel likelihood-free objective function that optimizes the mutual information between joint representations and the modalities that we are interested in generating at test time. Our proposed CMMD model furthermore uses an informative prior distribution that is conditioned on the modalities that are always available.

The empirical results show that the objective function we propose achieves higher downstream classification performance and lower rmse in the generated modalities than the regular variational lower bound. The model we propose also successfully counteracts the posterior collapse problem by optimizing mutual information, and by using an informative prior. Finally, the higher performance of our proposed CMMD model with respect to the state-of-the-art is consistent across different representative multi-modal problems. 

\addcontentsline{toc}{section}{Appendices}
\renewcommand{\thesubsection}{\Alph{subsection}}

\section*{Appendices}
\subsection{Objective Function}\label{ap_lowerbound}
\renewcommand{\theequation}{A\arabic{equation}}
\setcounter{equation}{0}
\renewcommand{\thetable}{C\arabic{table}}
\setcounter{table}{0}
\renewcommand{\thefigure}{C\arabic{figure}}
\setcounter{figure}{0}
\review{The joint distribution in the CMMD model is $p(\x_{\smallM},y,\z|\x_{\smallO}) = p(\x_{\smallM}|\x_{\smallO},\z)p(y|\z)p(\z|\x_{\smallO})$ and, under this model specification, the posterior distribution $p(\z|\x_{\smallO},\x_{\smallM},y)$ is intractable. Therefore, CMMD uses VI and approximates the true posterior distribution with a variational density $q(\z|\x_{\smallO},\x_{\smallM},y)$. Hence, the variational lower bound on the marginal log-likelihood of a single observation is}
\begin{align}
\label{eq_appendix_elbo}    
    \log p(\x_{\smallM},y|\x_{\smallO}) =& \log \int p(\x_{\smallM},y,\z|\x_{\smallO}) d\z \nonumber \\
                      =& \log \int q(\z|\x_{\smallO},\x_{\smallM},y)\frac{p(\x_{\smallM},y,\z|\x_{\smallO})}{q(\z|\x_{\smallO},\x_{\smallM},y)} d\z \nonumber \\
                      =& \log \mathbb{E}_{q(\z|\x_{\smallO},\x_{\smallM},y)} \frac{p(\x_{\smallM},y,\z|\x_{\smallO})}{q(\z|\x_{\smallO},\x_{\smallM},y)} \nonumber \\
                      \geq&  \mathbb{E}_{q(\z|\x_{\smallO},\x_{\smallM},y)}\bigg[\log \frac{p(\x_{\smallM},y,\z|\x_{\smallO})}{q(\z|\x_{\smallO},\x_{\smallM},y)}\bigg] \nonumber \\
                      =& \mathbb{E}_{q(\z|\x_{\smallO},\x_{\smallM},y)}[\log p(\x_{\smallM}|\x_{\smallO},\z) + \log p(y|\z) + \log p(\z|\x_{\smallO}) \nonumber \\ 
                      -& \log q(\z|\x_{\smallO},\x_{\smallM},y)],
\end{align}
\review{where the inequality is a result of the concavity of log and Jensen's inequality.}

Now we can write the conditional mutual information term $I_e(\x_{\smallM},\z|\x_{\smallO})$ (which depends on the functional form of the encoder as denoted by the subscript), as follows 
\begin{align}
    I_e(\x_{\smallM},\z|\x_{\smallO}) =&  \mathbb{E}_{p(\x_{\smallO},\x_{\smallM},\z)}\Bigg[\log \frac{p_e(\x_{\smallM},\z|\x_{\smallO})}{p_e(\x_{\smallM}|\x_{\smallO})p_e(\z|\x_{\smallO})} \Bigg] \nonumber \\
    =& \mathbb{E}_{p(\x_{\smallO},\x_{\smallM},\z)} \Bigg[\log \frac{p_e(z|\x_{\smallO},\x_{\smallM})p_e(\x_{\smallM}|\x_{\smallO})}{p_e(\x_{\smallM}|\x_{\smallO})p_e(\z|\x_{\smallO})} \Bigg]\nonumber \\
    =& \mathbb{E}_{p(\x_{\smallO},\x_{\smallM},\z)} [\log p_e(z|\x_{\smallO},\x_{\smallM}) -\log p_e(\z|\x_{\smallO}) 
    +\log p(\z|\x_{\smallO}) - \log p(\z|\x_{\smallO})] \nonumber \\
    =& \mathbb{E}_{p(\x_{\smallM},\x_{\smallO})}\big[KL[p_e(z|\x_{\smallO},\x_{\smallM}) ||p(\z|\x_{\smallO})]\big] - \mathbb{E}_{p(\x_{\smallO})}\big[KL[p_e(\z|\x_{\smallO})||p(\z|\x_{\smallO})]\big],
    \label{eq_ap_mi}
\end{align}
where $p(\x_{\smallO},\x_{\smallM},\z) = p(\x_{\smallO})p(\x_{\smallM}|\x_{\smallO})p(\z|\x_{\smallO})$, $p(\x_{\smallM},\z|\x_{\smallO})=$ $p(\z|\x_{\smallM},\x_{\smallO})p(\x_{\smallM}|\x_{\smallO})$, $\int p(\z|\x_{\smallM},\x_{\smallO})p(\x_{\smallM})d\x_{\smallM} \approx 1/N \sum_n p_e(\z|\x_{\smallO},\x_{\smallM}^n) = p_e(\z|\x_{\smallO})$, and all probability density functions are approximated by  variational approximations (the encoder and prior distribution in our proposed model). The  expectations $\mathbb{E}_{p(\x_{\smallM},\x_{\smallO})}$ and $\mathbb{E}_{p(\x_{\smallO})}$ are finally estimated using the empirical data distribution $\tilde{p}_D$.

Adding the conditional mutual information term $(1-\omega) I_e(\x_{\smallM},\z|\x_{\smallO})$ to the lower bound in Eq. \ref{eq_appendix_elbo} (mutual information optimization being controlled by $\omega \in [0,1]$) and replacing $p_e$ with the encoder $q(\z|\x_{\smallO},\x_{\smallM},y)$\footnote{\review{In this case the encoder is a variational approximation that can take any arbitrary form as long as it is a valid probability distribution \citep{sutter2021generalized}.}} gives the likelihood-free objective function for a single data point
\begin{align}
&\mathcal{L}(\x_{\smallO},\x_{\smallM},y) + (1-\omega) I_e(\x_{\smallM},\z|\x_{\smallO})  \nonumber \\
=&\mathbb{E}_{q(\z|\x_{\smallO},\x_{\smallM},y)}[\log p(\x_{\smallM}|\x_{\smallO},\z) + \log p(y|\z) + \log p(\z|\x_{\smallO})  
-  \log q(\z|\x_{\smallO},\x_{\smallM},y)] \nonumber \\
+& (1-\omega) KL[q(z|\x_{\smallO},\x_{\smallM},y) ||p(\z|\x_{\smallO})] - (1-\omega) KL[q(\z|\x_{\smallO})||p(\z|\x_{\smallO})] \nonumber \\
=& \mathbb{E}_{q{(\z|\x_{\smallO},\x_{\smallM},y)}}[\log p(\x_{\smallM}|\x_{\smallO},\z)+ \log p(y|\z)] - \omega KL[q(\z|\x_{\smallO},\x_{\smallM},y)||p(\z|\x_{\smallO})] \nonumber \\
-& (1-\omega) KL[q(\z|\x_{\smallO})||p(\z|\x_{\smallO})] \nonumber \\
\equiv& \mathcal{J}(\x_{\smallO},\x_{\smallM},y).
\end{align}

Note that we can obtain unbiased samples from $q(\z|\x_{\smallO})$ by first randomly sampling tuples $(\x_{\smallM},y) \sim \Tilde{p}_D$ and then $\z \sim q(\z|\x_{\smallO},\x_{\smallM},y)$. These are used to estimate the MMD divergence term in Eq. \ref{eq_convex_elbo}. 

\subsection{Upperbound on Mutual Information}\label{ap_up_mi}
Using the last line in Eq. \ref{eq_ap_mi} and replacing $p_e$ with the encoder $q(\z|\x_{\smallO},\x_{\smallM},y)$, which acknowledges the access to a labeled data set, it follows that
\begin{equation}
\mathbb{E}_{p(\x_{\smallM},\x_{\smallO},y)}[KL[q(z|\x_{\smallO},\x_{\smallM},y) ||p(\z|\x_{\smallO})]] \geq I(\x_{\smallM},\z|\x_{\smallO})
\end{equation}
given that the KL divergence is strictly positive. The expectation can be estimated using the empirical data distribution $\tilde{p}_D$.

\begin{figure}[t!]
    \centering
    \includegraphics[scale=0.65]{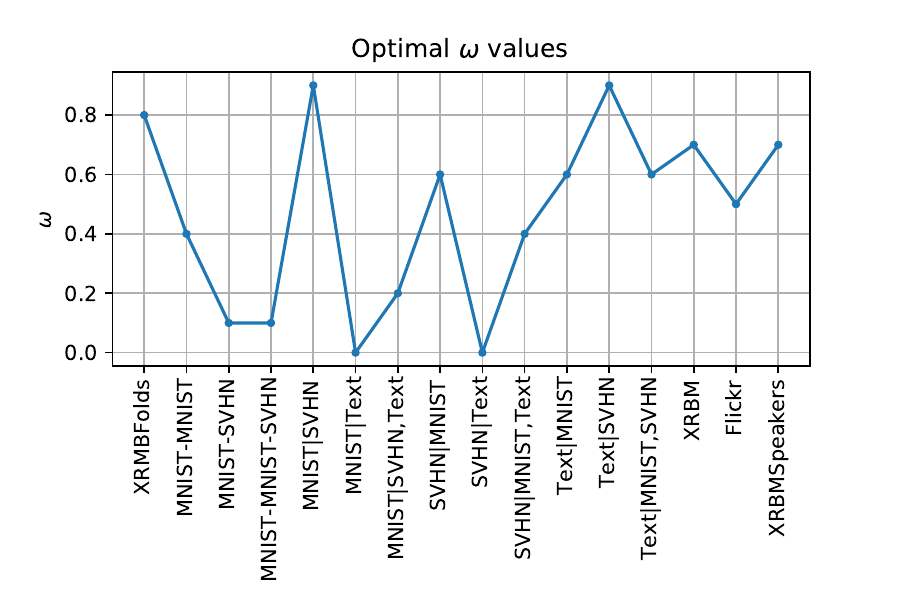}
    \caption{\review{Optimal $\omega$ value, found by cross-validation, for each of the experiments in this research. Experiments are ordered chronologically.}}
    \label{app_all_omegas}
\end{figure}

\subsection{Model Training and Architectures}
We minimized Eq. \ref{eq_convex_elbo} using SVGB and automatic differentiation routines in Theano \citep{team2016theano}. Note that the reconstruction term of Eq. \ref{eq_convex_elbo} can be efficiently estimated using the \textit{reparameterization trick} \citep{kingma2013auto}. The KL divergence term has a closed-form expression \citep{kingma2013auto, mancisidor2019deep}, and the MMD divergence is approximated numerically by drawing samples, as explained in Section A. This is the method suggested by \cite{zhao2017infovae} and \cite{rezaabad2020learning}.

CMMD architectures are, to provide a fair comparison in all experiments, chosen to resemble previous works. We furthermore use softplus activation functions in all hidden layers, using dropout \citep{srivastava2014dropout} with 0.2 probability. We use the same $\alpha$ and $\lambda$ parameter values for all CMMD models, which are set to 10 and 1000 respectively. We furthermore tune the hyperparameter $\omega$ over the grid $[0,0.1,0.2,\cdots,1]$. \review{Figure \ref{app_all_omegas} shows the optimal value of $\omega$ for each experiment in this research, which is found by cross-validation}. Finally, we use the Adam optimizer \citep{kingma2014adam} with a  $10^{-4}$ learning rate in all experiments. Our model is implemented in Theano and trained on a GeForce GTX 1080 GPU.

\textbf{Image-to-Image with MNIST}: 
The encoder network uses 3 hidden layers of 2500 neurons. Both the prior distribution and the decoder use 3 layers of 1024 neurons. The latent variable is a 50D vector and the classifier uses 2 hidden layers of 50 neurons. We assume, given that the second view is almost a continuous variable, that it is Gaussian distributed.  

\textbf{Image-to-Image with MNIST-SVHN}: The encoder, decoder and prior distribution in this experiment have 1 hidden layer of 400 neurons. The latent representation is a 20D vector and the classifier has 2 hidden layers of 50 neurons each.

\textbf{3-modality MNIST}: We use the same encoder in this experiment as in the "image-to-image with MNIST" experiment. The decoder architecture is shown in Table \ref{tbl_svhn_layers} (Decoder columns), which is the same architecture as in \cite{shi2019variational}. We add an extra layer, such as the one at the bottom of Table \ref{tbl_svhn_layers}, but with 1 stride to generate two missing modalities ($\x_2$ and $\x_3$). Note that we, for the rotated MNIST images, pad the images to a 32 $\times$ 32 matrix during training, and crop-back to a 28 $\times$ 28 matrix at test time. The decoder loss is, finally, the sum of two cross entropy terms, one for each missing modality. 

\textbf{Acoustic-to-Articulatory with XRMB}: 
We trained our model using the same 35 speakers  used by \cite{wang2015deep,wang2016deep}. The current version of the test data set, however, only contains 8 speakers without silence frames (silence frames were removed in the other 35 speakers). Our model is, for this reason, tested on 8 speakers in a speaker-independent downstream classification task (Table \ref{table1}). 

We use an encoder with 3 hidden layers of 3000 neurons. The prior distribution and decoder each have 3 hidden layers of 1500 neurons. The classifier model has 2 hidden layers of 100 neurons and the latent shared representation is a 70D vector. We assume a Gaussian distribution for modality $\x_{\smallM}$ in this case. The $\omega$ parameter has, for this data set, a significant impact on downstream classification and our best model uses $\omega=0.7$.

\textbf{Image-to-Annotation with Flickr}: 
We use an encoder with 4 hidden layers of 2048 neurons each. The prior distribution and decoder use 4 hidden layers of 1024 neurons. We, given that the modality $\x_{\smallM}$ corresponds to tags, use a Bernoulli decoder. The shared representation is a 1024D vector and our best model uses $\omega=0.5$. We deal with multi-label classification in this data set. The classifier for this model therefore uses 2000 neurons with sigmoid activations in the output layer and 2 hidden layers of 1550 neurons. Note that we follow previous works and exclude unlabeled images if they have less than 2 tags, given that we are interested in finding joint representations for both data-modalities. We finally standardize all features in the modality $\x_{\smallO}$.

\subsection{Generating Tags}
Table \ref{table_1a} shows  tags generated using our proposed CMMD model for some labeled images in the Flickr data set. Note that none of the images have any tag in the original data set.
\begin{table*}[t]
\centering
    \scriptsize 
    \begin{tabular}{c|ccccc}
                        & \includegraphics[height=1.3cm,keepaspectratio]{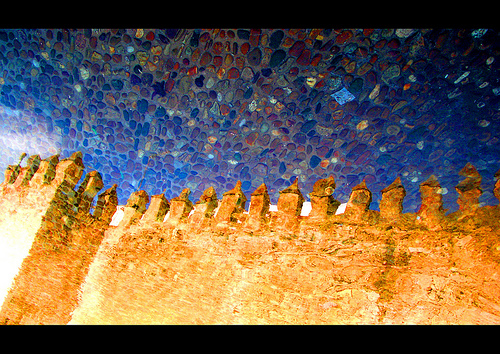} 
                        & \includegraphics[height=1.3cm,keepaspectratio]{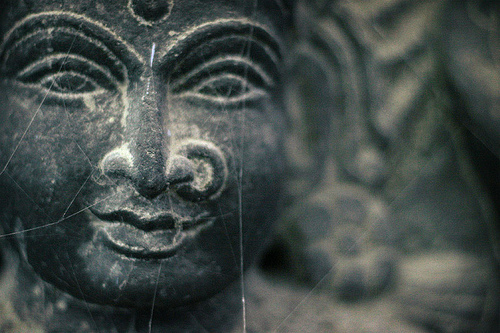}  
                        & \includegraphics[height=1.3cm,keepaspectratio]{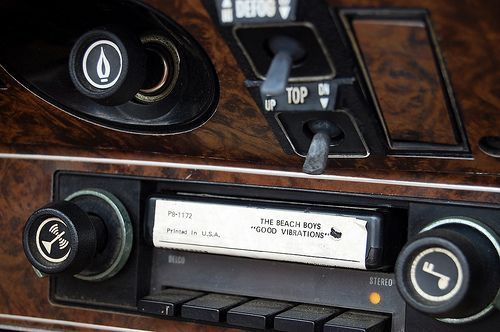}  
                        & \includegraphics[height=1.3cm,keepaspectratio]{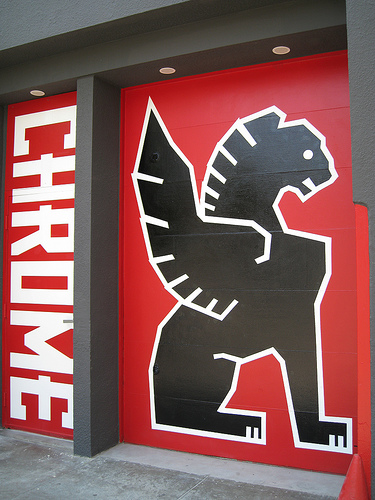}  
                        & \includegraphics[height=1.3cm,keepaspectratio]{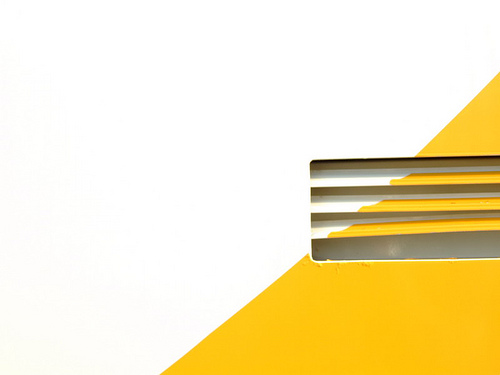}  
                        \\
        \hline
        Generated       & reflection, 
                        & beauty, stone,
                        & car
                        & chrome
                        & white, yellow,
                        \\
        tags            & themoulinrouge
                        & sculpture
                        & 
                        & 
                        & abstract, bus,
                        \\
                        & 
                        & 
                        & 
                        & 
                        & lines, graphic
                        \\
        \hline
                        & \includegraphics[height=1.3cm,keepaspectratio]{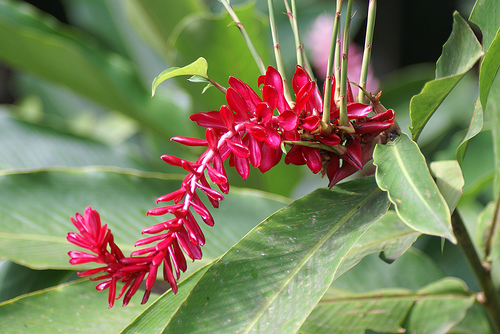}  
                        & \includegraphics[height=1.3cm,keepaspectratio]{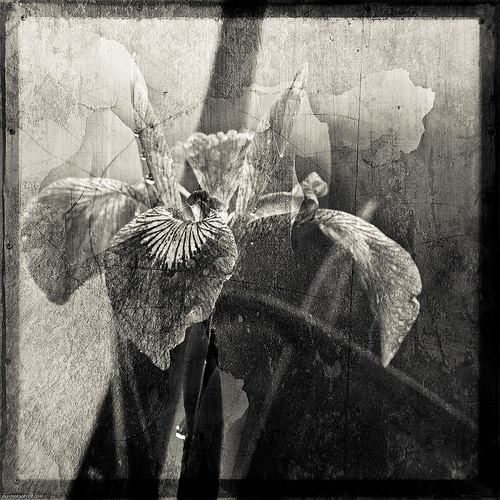}  
                        & \includegraphics[height=1.3cm,keepaspectratio]{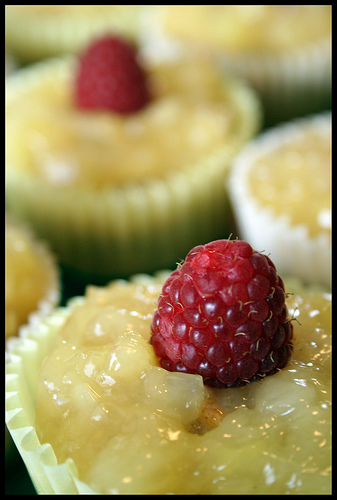}  
                        & \includegraphics[height=1.3cm,keepaspectratio]{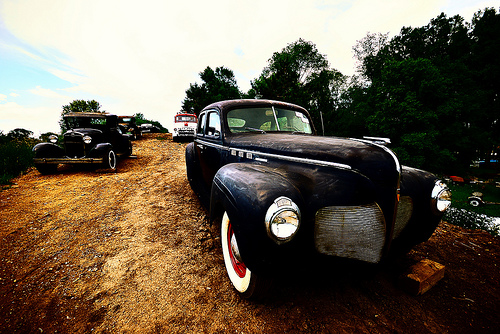}  
                        & \includegraphics[height=1.3cm,keepaspectratio]{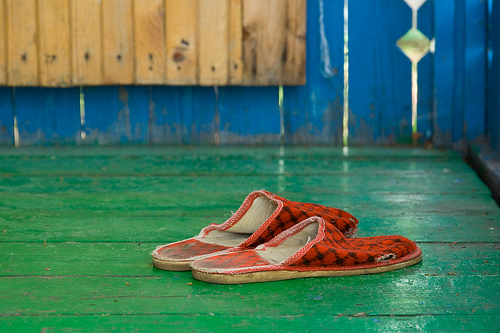}  
                        \\
        \hline
        Generated      & flower, holiday,
                        & flower, layers,
                        & food, vegan,
                        & d80
                        & landscape, 
                        \\
        tags            & vacation, red
                        & textures, iris,
                        & cupcake
                        & 
                        & explore,  
                        \\
                        & 
                        & soe
                        & 
                        & 
                        & flickr 
                        \\
        \hline
                        & \includegraphics[height=1.3cm,keepaspectratio]{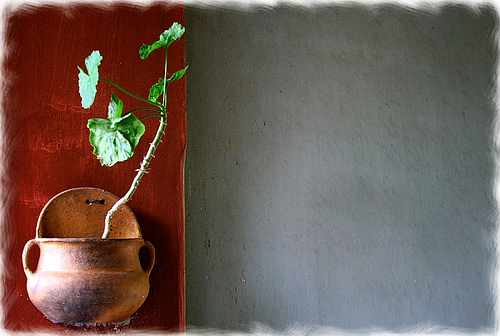}  
                        & \includegraphics[height=1.3cm,keepaspectratio]{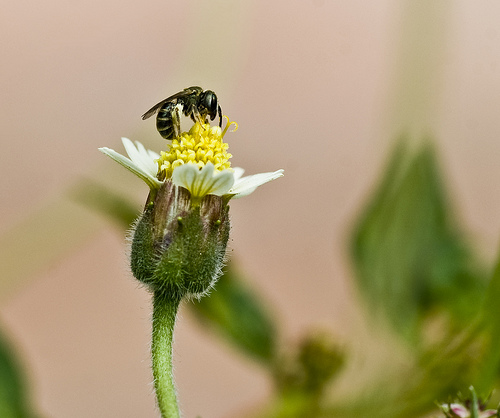}  
                        & \includegraphics[height=1.3cm,keepaspectratio]{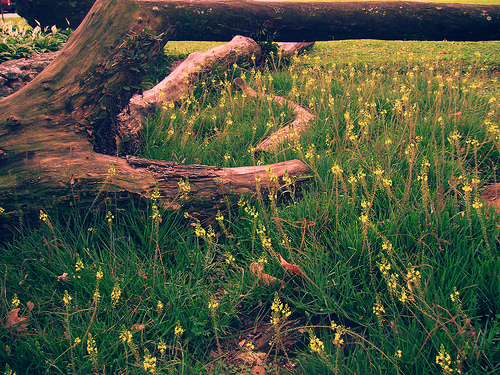}  
                        & \includegraphics[height=1.3cm,keepaspectratio]{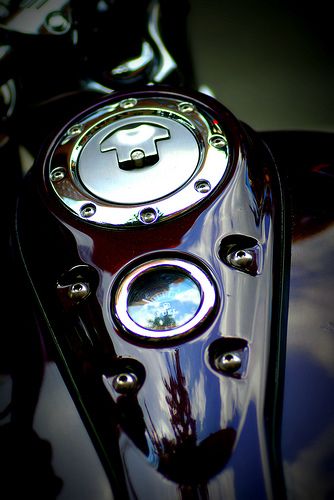}  
                        & \includegraphics[height=1.3cm,keepaspectratio]{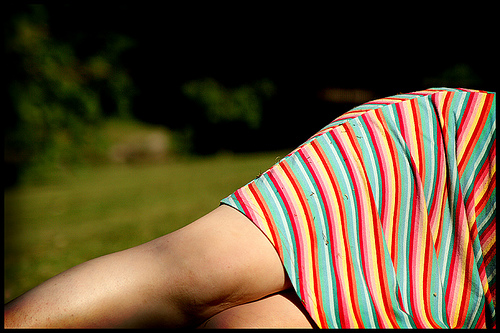}  
                        \\
        \hline
        Generated       & abigfave, minimal,
                        & macro, tamron,
                        & flower, explore,
                        & bike, 
                        & stripes
                        \\
        tags            & buenosaires, wall,
                        & nikond40, india
                        & flores
                        & red, tree, nyc,
                        & 
                        \\
                        & impressedbeauty
                        & 
                        & 
                        & ny, 
                        & 
                        \\
            \hline
                        & \includegraphics[height=1.3cm,keepaspectratio]{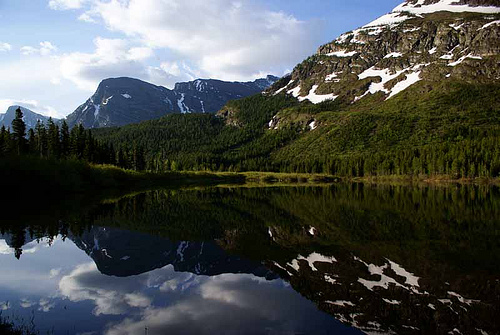}  
                        & \includegraphics[height=1.3cm,keepaspectratio]{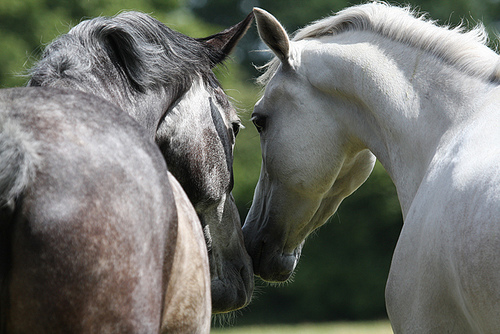}  
                        & \includegraphics[height=1.3cm,keepaspectratio]{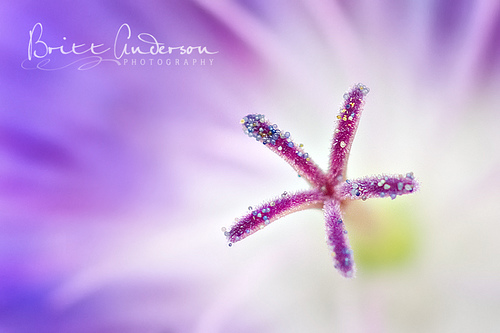}  
                        & \includegraphics[height=1.3cm,keepaspectratio]{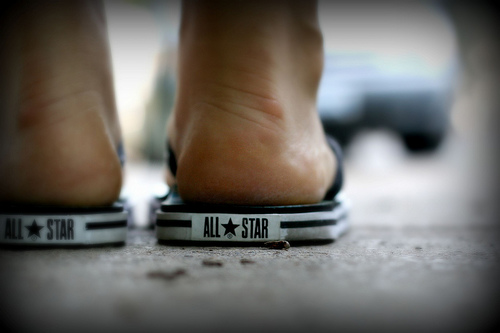}  
                        & \includegraphics[height=1.3cm,keepaspectratio]{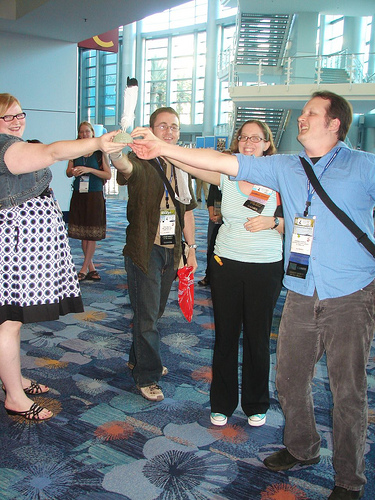}  
                        \\
        \hline
        Generated       & nature, water
                        & horses, grey,
                        & macro, garden,
                        & shoes, explore, 
                        & california
                        \\
        tags            & abigfave, landscape,
                        & friends
                        & closeup
                        & selfportrait, 
                        & 
                        \\
                        & reflection, mirror
                        & 
                        & 
                        & 365days, toronto
                        & 
                        \\
    \end{tabular}
    \caption{Tags generated using our proposed CMMD for some labeled images in the Flickr data set.}
    \label{table_1a}
\end{table*}

\subsection{Additional Details on Posterior Collapse}
We use the posterior collapse definition introduced in \cite{lucas_dont_2019}. This, in our experiments, is $Pr(KL[q(\cdot)||p(\cdot)] < \epsilon) \geq 1-\delta$, where $\delta = 0.01$ and $\epsilon \in [0,6]$. We therefore measure the proportion of latent dimensions \textit{i} that are within $\epsilon$ KL divergence for at least $1 - \delta$ of the data points. The MMVAE, MVAE, and VCCA models are, in our experiments, trained using the authors' publicly available codes\footnote{MMVAE: \url{https://github.com/iffsid/mmvae}, \\ MVAE: \url{https://github.com/mhw32/multimodal-vae-public},\\ VCCA: \url{https://ttic.uchicago.edu/~wwang5/}}. 

Fig. \ref{fig_collapse} shows different measures of collapse for Fold 1\footnote{The other 3 folds show the same pattern.} for the  section \ref{sec_vimodels} experiments. The far left diagram shows posterior collapse $Pr(KL[(z_i|\x_{\smallO})||(z_i)] < \epsilon) \geq 1-\delta)$, where $(\z_i) \sim \mathcal{N}(0,1)$ and $(z_i|\x_{\smallO})$ are drawn from the prior distribution, the joint MoE posterior, the joint PoE posterior, and the shared inference distribution for CMMD, MMVAE, MVAE, and VCCA respectively. 

The second diagram in Fig. \ref{fig_collapse} calculates $Pr(KL[(z_i|\x_{\smallM})||(z_i)] < \epsilon) \geq 1-\delta)$. However, $z_i \sim N(0,1)$ and $(z_i|\x_{\smallM})$ are, for this, drawn from the inference posterior distribution, the joint MoE posterior, the joint PoE posterior, and the inference private distribution in CMMD, MMVAE, MVAE, and VCCA respectively. Finally, the third diagram in Fig. \ref{fig_collapse} calculates $Pr(KL[(z_i|\x_{\smallO})||(z_i|\x_{\smallM})] < \epsilon) \geq 1-\delta)$, where $(z_i|\x_{\smallO})$ and $(z_i|\x_{\smallM})$ are drawn as explained above.

\subsection{Latent Space - MNIST}
\renewcommand{\thefigure}{F\arabic{figure}}
\setcounter{figure}{0}
\begin{figure}[t!]
  \centering
  \subcaptionbox*{}{\includegraphics[scale=.4]{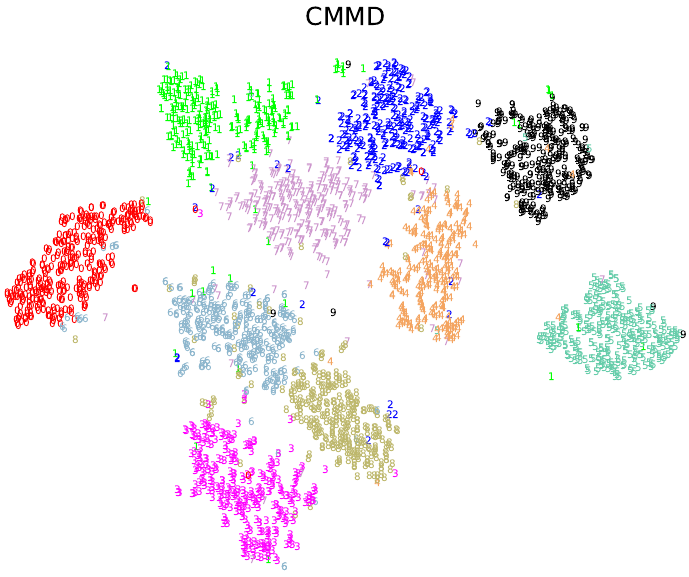}}
  \subcaptionbox*{}{\includegraphics[scale=.4]{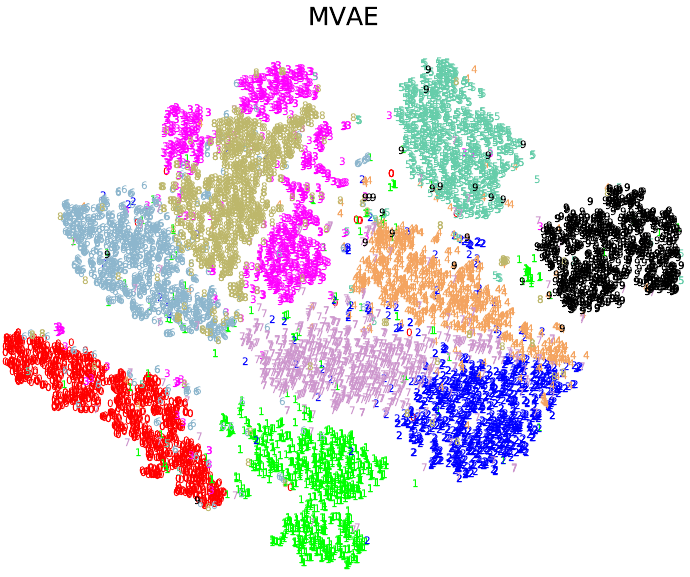}}
  \subcaptionbox*{}{\includegraphics[scale=.35]{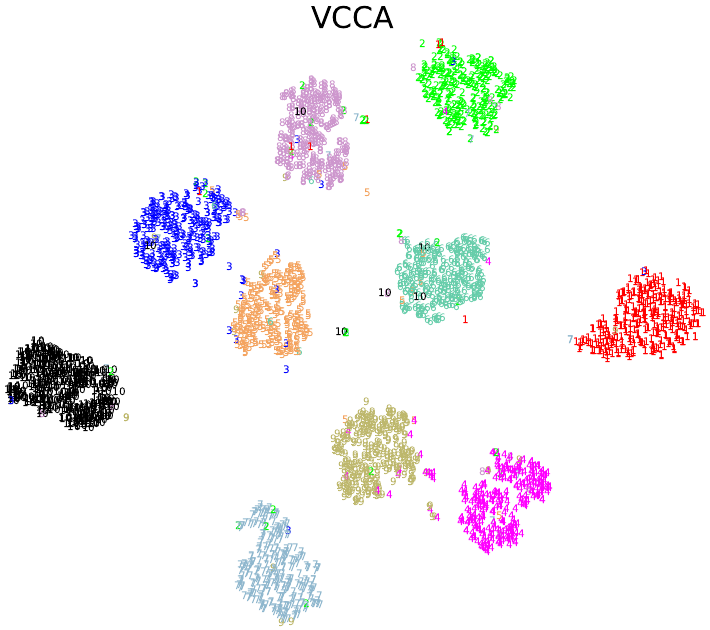}}
  \vspace{-1em}
  \caption{2D t-SNEs of the latent space in CMMD, MVAE and VCCA. The scatter color is assigned by the class label.}
  \label{fig_latent}
  \vspace{-0em}
\end{figure}
Figure \ref{fig_latent} shows 2D t-SNEs \citep{van2008visualizing} of the latent space learned using CMMD, MVAE and VCCA. The t-SNEs for both CMMD and VCCA show well separated class labels. Note that the class label variability is larger for the CMMD embeddings than VCCA. The t-SNEs for MVAE, however, show some overlapping class labels. 

\subsection{Generating Multiple Missing Modalities}
\renewcommand{\thefigure}{G\arabic{figure}}
\setcounter{figure}{0}
\begin{figure}[t!]
  \centering
  \subcaptionbox*{(a) CMMD - 1 missing modality}{\includegraphics[width=6cm,keepaspectratio]{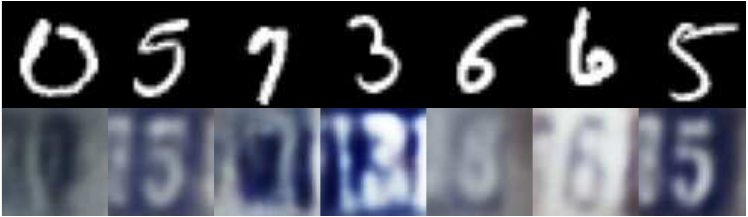}}
  \subcaptionbox*{(b) CMMD - 2 missing modalities}{\includegraphics[width=6cm,keepaspectratio]{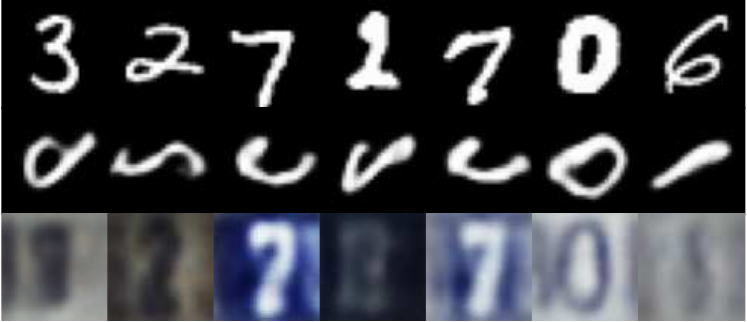}}\\
  \subcaptionbox*{(c) MMVAE-ELBO}{\includegraphics[width=6cm,keepaspectratio]{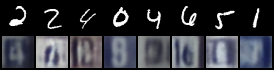}}
  \subcaptionbox*{(d) MMVAE-IWAE}{\includegraphics[width=6cm,keepaspectratio]{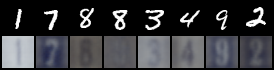}}\\
  \subcaptionbox*{(e) MVAE}{\includegraphics[width=6cm,keepaspectratio]{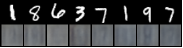}}
  \vspace{-0em}
  \caption{Generated images using CMMD (top row), MMVAE (middle row), and MVAE (bottom row). We, for all models, use the original MNIST digits to draw latent representations, which are further used to generate SVHN digits. Note that the MVAE images are taken from \cite{shi2019variational}. }
  \label{fig_recons}
  \vspace{-1em}
\end{figure}
We, in this section, compare the  missing modality/modalities generated at test time by the decoders in CMMD, MMVAE, and MVAE. In panel (a) we assume SVHN digits are missing at test time, in panel (b) both rotated-MNIST and SVHN are missing modalities at test time. In panel (c) and (d) we train MMVAE, optimizing the evidence lower bound (ELBO) and its importance weighted autoencoder (IWAE) version respectively, and generate the missing modality at test time (SVHN digits). For completeness, panel (e) shows the SVHN digits generated using MVAE reported in \cite{shi2019variational}. Both CMMD (panel (a)) and MMVAE-IWAE (panel (d)) generate quality and coherent SVHN digits, matching the MNIST digit in all cases. MVAE (panel (e)), however, generates low quality SVHN digits and it is difficult to see whether the generated image matches the MNIST digit. CMMD generates two missing modalities in panel (b), which is clearly a more challenging task. Only digits 7 and 0 are generated correctly for both missing modalities. Finally, it is interesting to compare the results obtained with MMVAE using two objective functions. If MMVAE optimizes the evidence lower bound, then the generated SVHN images have relatively low quality and do not match the MNIST class. 

\subsection{Cross-modal Generation with MNIST-SVHN-Text}
\renewcommand{\thetable}{H\arabic{table}}
\setcounter{table}{0}
\review{The network architectures used in the experiments using the MNIST-SVHN-Text data set are shown in Table \ref{tbl_svhn_layers}, \ref{tbl_mnist_layers}, and \ref{tbl_text_layers}, which are the same architectures used in \cite{shi2019variational,sutter2020multimodal,sutter2021generalized} and \cite{javaloy2022mitigating}. The only difference is that 
we use the encoder architecture in the aforementioned methods for the prior distribution in the CMMD model. The encoder architecture in the CMMD model is a fully-connected neural network with 3 hidden layers, each with 2500 units. All layers in the encoder use softplus activation functions and a dropout layer with 0.2 probability. Following previous work, the multimodal representation is a 20D latent variable, and we use the same values for $\alpha$ and $\lambda$ as in the other experiments, which are 10 and 1000 respectively. It is noteworthy that the 3 modalities are vectorized and concatenated before sending them through the encoder.}
\begin{table}[t!]
\caption{\review{Accuracy performance, averaged over 5 different runs, for all subsets of observable modalities. We do not include results for the method introduced in \cite{javaloy2022mitigating} given that the authors only provided average values over the different set of observable modalities (see Table \ref{tbl_allelbos}).}}
\centering
\def\arraystretch{0.95}
\setlength{\tabcolsep}{3.pt}
\footnotesize
\begin{tabular}{|cccccccc|}
\hline
 Model   & M    & S &   T   &   M,S    & M,T   &   S,T &   M,S,T \\
\hline
 MVAE   & $0.90 \pm 0.01$ & $0.44 \pm 0.01$ & $0.85 \pm 0.10 $ & $0.89 \pm 0.01$ & $0.97 \pm 0.02$ & $0.81 \pm 0.09$ & $0.96 \pm 0.02$ \\
 MMVAE  & $0.95 \pm 0.01$ & $0.79 \pm 0.05$ & $0.99 \pm 0.01 $ & $0.87 \pm 0.03$ & $0.93 \pm 0.03$ & $0.84 \pm 0.04$ & $0.86 \pm 0.03$ \\
 MoPoE  & $0.95 \pm 0.01$ & $0.80 \pm 0.03$ & $0.99 \pm 0.01 $ & $0.97 \pm 0.01$ & $0.98 \pm 0.01$ & $0.99 \pm 0.01$ & $0.98 \pm 0.03$ \\
 CMMD   & $0.98 \pm  \text{4E-3}$ & $0.80 \pm  \text{4E-3}$ & $1.00 \pm 0.00 $ & $0.99 \pm \text{2E-3}$ & $1.00 \pm 0.00$ & $1.00 \pm 0.00$ & $0.99 \pm  \text{2E-3}$ \\
\hline
\end{tabular}
\label{tbl_self-classification}
\end{table}
\begin{table}[t!]
\caption{\review{Accuracy values (generation coherence), averaged over 5 different runs, of the modalities conditionally generated by the CMMD model, together with the optimal $\omega$ values found by cross-validation.}}
\centering
\def\arraystretch{0.95}
\setlength{\tabcolsep}{10pt}
\footnotesize
\begin{tabular}{|c|ccc|ccc|ccc|}
\hline
    & \multicolumn{3}{c|}{M} & \multicolumn{3}{c|}{S} & \multicolumn{3}{c|}{T} \\
\cline{2-10}    
 & S & T & S,T & M & T & M,T & M & S &  M,S \\
\hline 
avg coherence & 0.75 & 1.00 & 1.00 & 0.66 & 0.87 & 0.87 & 0.98 & 0.69 & 0.98 \\
std coherence   & 0.02 & 0  &  0   &  0.07   & 0.07 &  0.10 &  \text{2E-3}   & 0.02    & \text{4E-3}   \\
$\omega$        & 0.9  & 0 & 0.2 & 0.6 & 0 & 0.4 & 0.6 & 0.9 & 0.6 \\
\hline
\end{tabular}
\label{tbl_omega_cross}
\end{table}

\review{To make a fair comparison with previous methods, we implemented a two-step classification using the multinomial logistic regression model implemented by scikit-learn with default values. The logistic regression model is trained using 500 latent variables, which are generated and randomly selected from the train data set. Finally, we test the predictive power of the trained logistic regression on the entire test data set. Table \ref{tbl_self-classification} shows the average accuracy over 5 different runs, for all subsets of observed modalities. For the cross-modal generation experiments, we train classifier models for each of the original unimodal modalities. The network architectures are the same as in the conditional prior column of Table \ref{tbl_svhn_layers}, \ref{tbl_mnist_layers}, and \ref{tbl_text_layers}, which again are the same architectures used in the aforementioned methods. Table \ref{tbl_omega_cross} shows the average coherence and standard deviation over 5 different runs, together with the optimal $\omega$ value found by cross-validation in each experiment.}

\begin{table}[t!]
\caption{\review{SVHN conditional prior and decoder layers. The last column for each model specifies the kernel size, stride, padding, and dilation. All layers are 2D convolutional (conv) and upconvolutional (upconv) in the encoder and decoder, respectively, with ReLU activations. Finally, the number of input and output dimensions in each layer is shown in the columns \#F.In and \#F.Out, respectively.}}
\centering
\def\arraystretch{0.95}
\setlength{\tabcolsep}{7.8pt}
\footnotesize
\begin{tabular}{lcccc|lcccc}
\hline
\multicolumn{5}{c}{Conditional prior} & \multicolumn{5}{c}{Decoder} \\
 Layer & Type & \#F.In & \#F.Out & Spec. & Layer & Type & \#F.In & \#F.Out & Spec. \\
\hline 
1 & conv    &   3   & 32    & (4,2,1,1)   & 1 & linear  & 20 & 128    &   \\
2 & conv    &   32   & 64    & (4,2,1,1)  & 2 & upconv  & 128 & 64    & (4,2,0,1)  \\
3 & conv    &   64   & 64    & (4,2,1,1)  & 3 & upconv  & 64 & 64    &  (4,2,1,1) \\
4 & conv    &   64   & 128    & (4,2,0,1) & 4 & upconv  & 64 & 32    &  (4,2,1,1) \\
5a & linear &   128   & 20    &           & 5 & upconv  & 32 & 3     &  (4,2,1,1) \\
5b & linear &   128   & 20    &           & & & & &   \\
\hline
\end{tabular}
\label{tbl_svhn_layers}
\end{table}

\begin{table}[t!]
\caption{\review{MNIST conditional prior and decoder layers. All layers are linear with ReLU activations. Finally, the number of input and output dimensions in each layer is shown in the columns \#F.In and \#F.Out, respectively.}}
\centering
\def\arraystretch{0.95}
\setlength{\tabcolsep}{14pt}
\footnotesize
\begin{tabular}{lccc|lccc}
\hline
\multicolumn{4}{c}{Conditional prior} & \multicolumn{4}{c}{Decoder} \\
 Layer & Type & \#F.In & \#F.Out & Layer & Type & \#F.In & \#F.Out \\
\hline 
1  & linear  &   784   & 400  & 1 & linear  & 20 & 400  \\
2a & linear  &   400   & 20   & 2 & linear  & 400 & 784 \\
2b & linear  &   400   & 20   &   & & &  \\
\hline
\end{tabular}
\label{tbl_mnist_layers}
\end{table}

\begin{table}[t!]
\caption{\review{Text conditional prior and decoder layers. The last column for each model specifies the kernel size, stride, padding, and dilation. All layers are 1D convolutional (conv) and upconvolutional (upconv) in the encoder and decoder, respectively, with ReLU activations. Finally, the number of input and output dimensions in each layer is shown in the columns \#F.In and \#F.Out, respectively.}}
\centering
\def\arraystretch{0.95}
\setlength{\tabcolsep}{7.8pt}
\footnotesize
\begin{tabular}{lcccc|lcccc}
\hline
\multicolumn{5}{c}{Conditional prior} & \multicolumn{5}{c}{Decoder} \\
 Layer & Type & \#F.In & \#F.Out & Spec. & Layer & Type & \#F.In & \#F.Out & Spec. \\
\hline 
1 & conv    &   71   & 128   & (1,1,0,1)    & 1 & linear  & 20 & 128    &   \\
2 & conv    &   128   & 128    & (4,2,1,1)  & 2 & upconv  & 128 & 128   & (4,1,0,1)  \\
3 & conv    &   128   & 128    & (4,2,0,1)  & 3 & upconv  & 128 & 128   & (4,2,1,1)  \\
4a & linear &   128   & 20    &             & 4 & conv    & 128 & 71     &  (1,1,0,1) \\
4b & linear &   128   & 20    &             & & & & &   \\
\hline
\end{tabular}
\label{tbl_text_layers}
\end{table}

\bibliography{bibliography}
\bibliographystyle{myapalike}

\end{document}